\pdfoutput=1

 \documentclass[10pt,twocolumn,twoside,journal]{IEEEtran}

\usepackage{cs}
\usepackage{mathsymb}
\usepackage{verbatim}
\usepackage{subfigure}
\usepackage{url}
\usepackage{graphicx}
\graphicspath{{./}{./Fig/}{./Figs/}{./Fig/eps/}{./Fig/pdf/}}

\usepackage{cite,eucal}

\usepackage{algorithm2e}
\let\savedalgorithm\algorithm
\let\savedendalgorithm\endalgorithm
\newenvironment{algorithmic}{%
\savedalgorithm
}{%
\savedendalgorithm
}

\usepackage{algorithm}

\usepackage{amsmath}
\usepackage{url}
\usepackage{multirow}

\def\Integer{\mathbb{Z}^+}
\def\I{{\bf I}}

\newcommand{\bigO}{\ensuremath{\mathcal{O}}}

\begin{document}

\title{Asymmetric Pruning For Learning \\  Cascade Detectors}

\author{
         Sakrapee Paisitkriangkrai,
         Chunhua Shen,
         Anton van den Hengel
\thanks
{
}
\thanks
{
The authors are with The Australian Center for Visual Technologies,
The University of Adelaide, SA 5005, Australia
(e-mail: \{paul.pais, chunhua.shen, anton.vandenhengel\}@adelaide.edu.au).
This work was in part supported by ARC Grant FT120100969.
Correspondence should be addressed to C. Shen.
}
\thanks
 {
 }
}

\markboth{Manuscript}
{Paisitkriangkrai
\MakeLowercase{\textit{et al.}}: Asymmetric Pruning for Learning
Cascade Detectors}

\maketitle

\begin{abstract}

Cascade classifiers are one of the most important contributions to
real-time object detection.  Nonetheless, there are many challenging
problems arising in training cascade detectors.  One common issue is
that the node classifier is trained with a symmetric classifier.
Having a low misclassification error rate does not guarantee an
optimal node learning goal in cascade classifiers,
\ie, an extremely high detection rate with a moderate false positive
rate.  In this work, we present a new approach to train an effective node
classifier in a cascade detector.  The algorithm is based on two key
observations:
1) Redundant weak classifiers can be safely discarded;
2) The final detector
should satisfy the asymmetric learning objective of the cascade
architecture.  To achieve this, we separate the classifier training
into two steps: finding a pool of discriminative  weak
classifiers/features and training the final classifier by pruning weak
classifiers which contribute little to the asymmetric learning
criterion (asymmetric classifier construction).  Our model reduction
approach helps accelerate the learning time while achieving the
pre-determined learning objective.  Experimental results on both face
and car data sets verify the effectiveness of the proposed algorithm.
On the FDDB face data sets, our approach achieves the state-of-the-art
performance, which demonstrates the advantage of our approach.

\end{abstract}


\section{Introduction}

    Real-time object detection is a fundamental topic in computer vision due to its tremendous uses
    in many applications such as video surveillance, real-time human computer interaction, robotics,
    \etc \cite{Viola2004Robust,Nascimento2006,Pallavi2008,Siddiquie2012}.
    The task of object detection is to identify predefined objects in a given image using
    knowledge learned from pre-labeled objects.  Among various real-time object detection
    algorithms, Viola and Jones' algorithm \cite{Viola2004Robust} is the most commonly adopted
    approach due to its effectiveness and efficiency.  Their framework consists of two phases.  The
    first phase discovers and learns discriminative features from a large set of feature pools
    (feature extraction).  Extracted features are used to construct a classifier in the second phase
    (classifier learning).  In \cite{Viola2004Robust}, the authors combined these two phases
    together through the use of AdaBoost.  AdaBoost selects relevant features and at the same time
    constructs a strong classifier.

    Significant effort has been spent on improving the Viola and Jones' framework.  One
    common technique is to post-adjust linear coefficients of weak classifiers selected by
    AdaBoost in order to introduce an asymmetric property into the cascade classifier for an
    effective rejection of negative patches in early nodes.  Post-processing algorithms can be
    divided into four categories:
(a) By tuning node thresholds during detector training, \eg, traditional cascade classifier
\cite{Viola2004Robust};
(b) By tuning node thresholds after the entire cascade classifier has been learned,
\eg, soft cascade \cite{Bourdev2005Robust},
optimized cascade \cite{Luo2005Optimization};
(c) By tuning weak classifiers and weak classifiers' coefficients during the cascade detector training,
\eg, the LAC classifier \cite{Wu2008Fast}.
(d) By tuning weak classifiers and weak classifiers' coefficients after
the entire cascade classifier has been learned,
\eg, the joint cascade \cite{Dundar2007Joint}.

\begin{figure}[t]
        \begin{center}
            \includegraphics[width=0.5\textwidth]{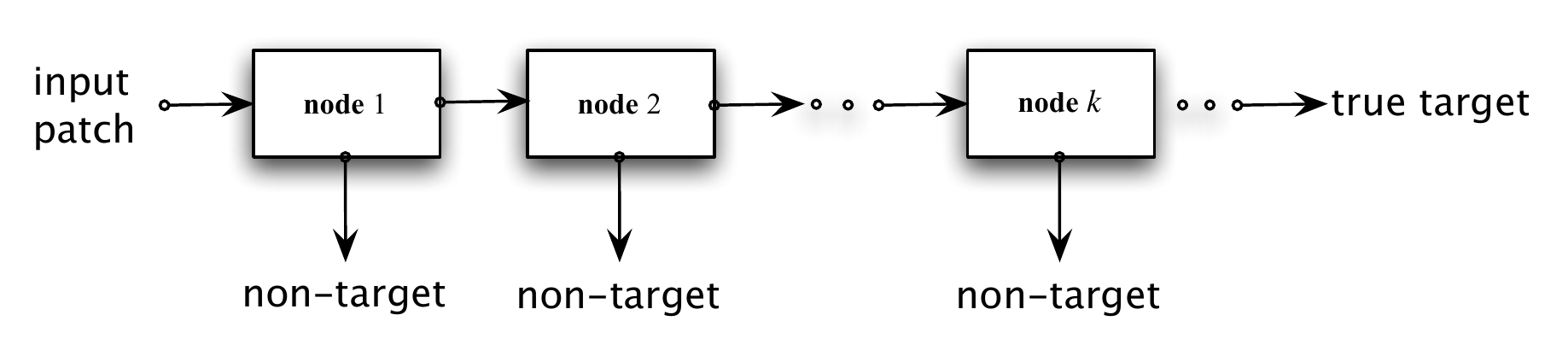}
        \end{center}
        \caption{The cascade classifier. The overall detection rate
        and false positive rate can be calculated using \eqref{EQ:0A}
        and \eqref{EQ:0B}. An input image patch is classified as a true
        detection only when it passes all the node classifiers.
        }
        \label{fig:cascade}
    \end{figure}

    A cascade classifier consists of a set of node classifiers
    (see Fig.\ \ref{fig:cascade} for an illustration). It
    is very different from a standard classifier in that
    the overall detection rate (classification accuracy on the positive data)
    can be approximately calculated as the product of the detection rate
    of each node classifier:
    \begin{equation}
        \label{EQ:0A}
        {\rm DR}_{\rm ovr} =  \prod_{k} {\rm DR}_k.
    \end{equation}
    The overall false positive rate (classification error rate
    on the negative data) is the product of the false positive rate of each node:
    \begin{equation}
        \label{EQ:0B}
        {\rm FP}_{\rm ovr} = \prod_k {\rm FP}_k.
    \end{equation}
    Here $ k $ indexes the node classifier.
    These two equations are valid under the assumption that each node makes
    independent classification errors.
    From these two equations, it is easy to see that
    in order to achieve a high overall detection rate and a low overall false positive rate,
    {\em each node classifier must achieve an extremely high detection rate and only
    a moderate false positive rate.}
    For instance, if the design goal for each node is to have a detection rate of
    $ 99.5\% $ and a false positive rate of around $ 50\% $ and the cascade classifier
    has $ 22 $ nodes in total, then the overall performance is:
        $ {\rm DR}_{ \rm ovr} \approx  90\%$  and
        $ {\rm FP}_{ \rm ovr} \approx  2 \cdot 10^{-7}  $.
        This is referred to as the {\em node learning goal} in \cite{Wu2008Fast}.

    In this work, we introduce a new post-processing approach by pruning AdaBoost's weak
    classifiers during the course of detector training.
    The intuition behind our pruning approach is to
    remove less discriminative weak classifiers while focusing on the asymmetric learning objective
    of cascade classifiers.
    In short, we use a fast asymmetric pruning technique to train
    the node classifier that better meets the node learning goal and consequently
    improves the overall performance of the cascade.

    Pruning is a well-known technique widely used in supervised
    learning such as feature selection.  It reduces the size of
    the model by removing model components that provide little or no discriminative power to
    classify instances.  It has been applied in a decision tree to remove nodes which are less
    significant \cite{Hastie2009Elements}.  By using pruning, one is able to reduce the complexity of the
    final classifier as well as achieve a better predictive accuracy.

    Boosting has been a method of choice for many learning problems including visual object
    detectors.  The algorithm constructs a strong classifier which consists of a linear combination
    of weak classifiers.  The training procedure of boosting is an iterative process.
    At each iteration, the algorithm selects the weak classifier which has minimal weighted error.
    Samples are reweighed based on its classification error in the previous round.
    The process continues until
    the maximum number of iterations is reached or no weak classifier can be added into the
    ensemble.
    One of the popular boosting algorithms is AdaBoost \cite{Freund1996Experiments}.
    Although AdaBoost has been commonly used in object detection, various researchers argue that
    AdaBoost is {\em sub-optimal} for training cascade classifiers \cite{Lienhart2003Empirical,
    Shen2011Efficiently, Wu2008Fast}.  A few alternative algorithms have been
    proposed to replace AdaBoost, \eg, asymmetric boosting
    \cite{PhamHC08,Wang2010ACCV,Viola02asymboost,wang2012fast,ICCV13Pai}
    and GentleBoost \cite{Lienhart2003Empirical}.

    Pruning ensemble classifiers obtained from AdaBoost is of interest for many reasons.  Firstly,
    AdaBoost is popular due to its simplicity and  efficiency.
    Weak classifiers' coefficients can be calculated in
    a closed form.  Although training a face detector was reported to be slow in the original Viola
    and Jones' framework (training a complete cascade classifier takes $4$ weeks), a
    recent study reveals that it is possible to speed up this training time by caching feature
    values at the start of the AdaBoost training \cite{Wu2008Fast}.  Secondly, AdaBoost is a well
    studied method and has been shown to be effective for many classification problems.  Thirdly, the
    final trained classifier is an ensemble classifier which is fast to compute during evaluation.
    Unfortunately, AdaBoost performs sub-optimally in terms of
    achieving the  asymmetric {\em node learning goal}
    \cite{Wu2008Fast}.  Furthermore, AdaBoost is an ensemble learning technique which uses a forward
    selection search strategy.  The algorithm is short-sighted and might not produce a near
    optimal classifier \cite{Shen2011Efficiently}.  Finally, AdaBoost reduces the training error
    rate by concentrating on examples that are difficult to classify.  As a result, AdaBoost may
    select irrelevant weak classifiers if the training data are noisy or contain outliers.

    In this work, we propose to prune weak classifiers trained by AdaBoost by eliminating less
    relevant features from the candidate feature pool.  To be more specific, we exclude weak
    classifiers which have a minimal impact on a class separation between positive and negative
    samples.  The criterion is not only capable of eliminating redundant features but also
    able to exploit
    the asymmetric node learning goal.  The resulting ensemble is a compact linear combination of
    weak classifiers which is fast to evaluate.  To perform pruning, we evaluate AdaBoost's weak
    classifiers using greedy sparse linear discriminant analysis (GSLDA)
    \cite{Moghaddam2006Generalized,Moghaddam2007Fast}.
    By combining AdaBoost
    and GSLDA, we are able to exploit the fast feature selection (via AdaBoost) and achieve the
    asymmetric node learning goal in the cascade architecture (via GSLDA).
    This is a novel application of GSLDA in real-time object detection.

    In summary, we use AdaBoost to select an over-complete weak classifier pool. At the second step,
    an asymmetric pruning method
    (here we use GSLDA) is then applied to remove less relevant weak classifiers against the
    asymmetric node learning criterion. In theory, other feature selection methods
    such as fast forward selection of Wu \etal \cite{Wu2003FFS} can be used
    to replace AdaBoost at the first step.

    The main contributions of the presented work can be summarized as follows.
\begin{enumerate}
  \item We propose an alternative method to train an ensemble classifier which leads to a further
      performance improvement.  The approach can be applied to
      many cascade classification based applications. The core of the proposed method
      is a novel application of the fast GSLDA algorithm.
  \item We apply pruning to two well-known frontal face detection data sets and car data sets.
      Better performance is observed over a few other cascade
      classifiers.
      On more challenging face data sets, the FDDB face data sets, our algorithm achieves
      state-of-the-art performance.
\end{enumerate}

    The rest of the paper is organized as follows.  Section~\ref{sec:related} briefly outlines
    related work on pruning and post-training in object detection.  Section~\ref{sec:approach}
    introduces background concepts of boosting and GSLDA.  We then propose our pruning approach to
    enhance object detection performance.  Experimental results are presented in
    Section~\ref{sec:exp}.  Finally, we conclude our paper in Section~\ref{sec:conc}.

\section{Related Work}
\label{sec:related}

    Over the past decade, the computer vision community has witnessed numerous
    success on real-time object
    detection.  Most of these work extended the original work of Viola and Jones'
    real-time face detector.  Viola and Jones' work consists of three major components:
1)
      The cascade classifier.  The cascade classifier can be represented as a degenerate tree.  It
      is designed to efficiently filter out negative patches in early nodes for real-time face
      detection.
2)
      AdaBoost. AdaBoost trains a strong classifier by selecting discriminative features from a pool
      of Haar-like features.
3)
      Integral images for fast computation of Haar-like features.

In the literature, there are a few approaches that attempt to improve the work of Viola and Jones.
In this section, we focus on those work which applies post-processing to the trained cascade
classifier.
By re-adjusting weak classifiers' coefficients and node thresholds, one can further improve the final performance of object detectors.
In the rest of this section, we review existing work related to pruning and post-training algorithms.

Zhang and Viola view the object detection problem as  multiple instance
learning  \cite{Zhang2007Multiple}.
They proposed to combine multiple instance pruning with soft cascade
\cite{Bourdev2005Robust}.  They first train a single boosted
classifier on the entire data set.
Instead of setting a node threshold using a simple heuristic rule,
they set the node
threshold so that at least one acceptable window will be retrained.
It is  demonstrated that
this post-training simplifies the training procedure and yields an improvement compared to the
traditional cascade classifier and soft cascade.
Chen and Chen proposed a novel cascaded structure called Meta-stages \cite{Chen2008Fast}.
The algorithm appends additional classifiers, termed meta-stages, to the original boosted cascade.
Meta-stages exploit information from previous nodes of the cascade classifier.
Li and Zhang argued that features selected by AdaBoost could be suboptimal since AdaBoost trains weak classifiers in a sequential forward selection manner \cite{Li2004Float}.
The authors introduced a boosting variant known as FloatBoost.
FloatBoost incorporates the idea of floating search into AdaBoost.
The algorithm backtracks and examines the already added weak
classifiers and discards the redundant ones.
They show that FloatBoost needs fewer weak classifiers than AdaBoost to achieve a similar performance.

Wu \etal argued that tuning the node threshold to achieve the asymmetric node learning goal is suboptimal for training the cascade classifier \cite{Wu2005ICML,Wu2008Fast}.
They proposed to decouple the problem of feature selection and ensemble classifier design in order to introduce asymmetry.
They proposed to use linear
asymmetric classifier (LAC) for post-processing weak learners'
coefficients. LAC is optimal  for the
node learning goal under the assumption that the linear projection of
negative samples' features is symmetric and the linear projection of
positive samples' features follows a Gaussian distribution.  LAC
maximizes the detection rate while discarding $50\%$ of negative
samples.
This objective criterion has proven to be effective for training the cascade classifier as later
shown in \cite{Shen2011Efficiently}.  Wu \etal observed that in some cases, linear discriminant
analysis (LDA) gives a better performance than LAC on face data sets.
Shen \etal proposed greedy sparse LDA (GSLDA) as an alternative
approach to train object detectors \cite{Shen2011Efficiently}.
They generate a set of discriminative features by training one weak classifier for each Haar-like features.
GSLDA is applied to sequentially select best weak classifiers.
The best weak classifier is the one that yields a maximal class separation when added to the current set.
The major drawback of their technique is that the algorithm trains only one weak classifier feature
for each Haar-like feature, similar to \cite{Wu2003FFS}.
Hence, there is room to improve their detection performance since the number of available
features is limited.
It is the work of \cite{Wu2005ICML,Wu2008Fast,Shen2011Efficiently} that has directly inspired
our work here.

Our pruning approach is different from existing approaches.
Unlike FloatBoost \cite{Li2004Float}, where less discriminative weak classifiers are sequentially
removed at each boosting iteration (floating search), we perform backward elimination with the
asymmetric node learning goal after we have completed training a boosted classifier.
Doing so not only results in a significantly reduced training time but also yields
a final classifier which satisfies the asymmetric node learning goal.
In contrast, FloatBoost does not take the asymmetric learning into account. The main purpose
of FloatBoost is only to remove redundant weak classifiers.

Our approach is also different from \cite{Wu2005ICML,Wu2008Fast}, where the authors
re-train the coefficients of weak classifiers learned from AdaBoost.
Since AdaBoost is greedy, \ie, choosing the weak classifier and the weak classifier's
coefficient in order to cause the greatest reduction in the exponential loss,
it can be short-sighted in choosing the best weak classifier at each iteration.
Hence the final set of weak classifiers, as used in \cite{Wu2005ICML,Wu2008Fast},
might not be optimal in order to achieve the node learning goal.
This can be observed in Fig. \ref{fig:singlenode}, where
the set of weak classifiers used in
\cite{Wu2005ICML,Wu2008Fast} is not optimal with
respect to fulfilling the asymmetric node
learning goal.

Compared to \cite{Shen2011Efficiently}, the size of our initial pool of weak
classifier features is not only smaller but also more discriminative than theirs.  Hence, the
training time is faster and the trained detector is more accurate.  In the next section, we present
the fast and effective pruning algorithm for training the visual object detector.

\section{The Approach}
\label{sec:approach}

    Our approach can be broken down into two steps.  In the first step, we perform a sequential
    forward search to learn a sufficiently large set of discriminative features.
In the second step, we perform a sequential backward elimination to construct a more compact binary
ensemble classifier.
In this section, we first discuss boosting based visual features.
We then briefly review the concept of post-training binary stump features with an asymmetric classifier.
Finally, we propose our pruning based feature selection algorithm.  For ease of exposition, symbols
and their denotations used in this section are summarized in Table~\ref{tab:notations}.

\begin{table}[t!]
  \caption{Notation}
  \centering
    \begin{tabular}{c|l}
    \hline
    Notation & Description \\
    \hline
    \hline
      $N$ & Number of training samples in each classifier\\
      $M$ & Number of low level features \\
      $D$ & Number of pixels for each training sample \\
      $L$ & Number of bins for histogram features \\
      $T$ & Number of features in the final classifier \\
      $T_1$ & Number of initial features to be selected \\
      $T_2$ & Number of features to be discarded during
             pruning ($T_1 - T$) \\
     \hline
    \end{tabular}
  \label{tab:notations}
\end{table}

    \subsection{Boosting based feature selection}
    Given a training data consisting of $N$ samples $\{(\bx_i,y_i)\}_{i=1}^N$
    where $\bx_i \in \mathbb{R}^M$ is a $M$ dimensional feature vector
    of the $i$-th sample and
    $y_i \in \{-1, 1\}$ denotes the class label of the $i$-th sample.
    Here any feature descriptors that map the original
    raw pixel features of $D$ dimensions to $M$ dimensions
    \eg, Haar-like features or SIFT features,
    can be applied.
    Our goal is to learn a prediction function that achieves
    the asymmetric node learning goal in the cascade classifier.
    In order to achieve this, we first transform the original $M$
    dimensional training data into another feature space, in which classes
    can be separated more easily.
    One possible transformation is to apply the $\sign(\cdot)$
    function to
    each input feature.
    The transformation function can be written as,
    \begin{equation}
    \label{EQ:1}
      h(\bx^j) = \sign \bigl( p ( \bx^j - \theta  ) \bigr), j \in [1,M],
    \end{equation}
    where $\bx^j$ is an input data at dimension $j$,
    $\theta$ is a threshold, $\theta \in [\min(\bx^j), \max(\bx^j)]$,
    and $p \in \{-1, 1\}$ is a polarity indicating the direction of the step function.
    In other words, we generate a classifier which partitions the input space into two sets: $p\bx^j < p \theta$ and $p\bx^j \geq p \theta$.
    Let us assume that each $M$ dimensional feature vector has $N$ distinct feature
    values,
    the maximum number of dimensions of the new feature space is $MN$.
    For vision tasks, \eg, face and car detection,
    the training sample in each node can be close to $10,000$ and
    the number of Haar-like features can be anywhere in the order of
    $10^6$, there would be more than one billion binary features to consider.
    Searching all possible subsets of features is computationally expensive and infeasible.
    Boosting can be used to select features in this extremely high
    dimension.

    Boosting is a well-known machine learning algorithm, which builds an additive model from a set of weak learners \cite{Schapire1999Boosting}.
    There are many variants of boosting, \eg, AdaBoost \cite{Freund1996Experiments}, LogitBoost \cite{Friedman2000Additive}, BrownBoost\cite{Freund2004Adaptive}, LPBoost \cite{Demiriz2002LPBoost}, \etc
    The algorithm trains series of weak learners with updated sample weights.
    The weak learning algorithm is designed to select the single feature which best separates positive and negative examples.
    For each feature, the weak learner determines the optimal classification function, such that a minimum number of examples are misclassified.
    A weak classifier consists of a feature, $f$, a threshold $\theta$, and a polarity, $p$, indicating the direction of the inequality,
    \begin{equation}
    \label{EQ:2}
    h(\bx, f, p, \theta) =
      \begin{cases} 1 & \text{if } p f(\bx) < p \theta \\
                    -1 & \text{otherwise, }
      \end{cases}
    \end{equation}
    where $\bx$ is a training example.
    The final decision rule is formed by linearly combining the set of hypotheses (weak learners) generated at each round with their weighted votes.
    The final prediction can be written as
    \begin{equation}
    \label{EQ:FxAda}
          F(\bx) = \sign \left( \sum_{t=1}^T \alpha_t h_t(\bx) \right),
    \end{equation}
    where $h_t$ is the $t$-th weak learner at iteration $t$ and $\alpha_t$ is the $t$-th coefficient
    computed by the boosting procedure.

    One of key advantages of applying boosting as a feature selection mechanism is the speed of
    learning.  In the traditional feature selection, the algorithm would need to evaluate $MN$
    binary weak classifiers (assume feature values are distinct).  On the other hand, boosting makes
    use of sample weights to compactly encode the dependency of previously selected features.  These
    weights can then be used to evaluate a given weak classifier in a constant time.  By applying
    boosting, we are able to efficiently select a subset of features, which are most discriminative
    for classification, from a very large pool of features.

    Boosting with decision stumps as weak classifiers combines two tasks simultaneously when
    training a classifier: selecting the subset of features and building the symmetric ensemble
    classifier.  For training the cascade classifier with the asymmetric learning objective (\eg,
    $99\%$ detection rate and $50\%$ false positive rate), separating these two processes provides
    more flexibility.
    In the next section, we briefly review the concept of linear asymmetric classifier
    (LAC), which has been shown to be a better alternative in learning an ensemble classifier for the cascade
    framework.

    \subsection{Post-processing with LAC}

    Wu \etal proposed LAC as a post-processing step for training nodes in the cascade framework
    \cite{Wu2008Fast}.
    They post-trained a weighted vote of AdaBoost's weak classifiers using the asymmetric
    criterion.
    In their work, one of the conclusions is that LAC is guaranteed to reach an optimal solution
    in terms of the node learning goal under
    the assumption of Gaussian data distribution.  In this section, we briefly review their
    approach, which motivates our proposed algorithm.

    Given a linear classifier $f( \x ) = \sign(\bw^\T \bx - b)$.
    The objective of each node in cascade classifiers is to seek a $\{\bw, b\}$ pair which has a very high accuracy on the positive data, $ \bx_1$, and moderate accuracy on the negative data, $\bx_2$.
    This objective can be expressed as the following optimization problem,
    \begin{align}
    \label{EQ:LAC_obj}
        \max_{ \bw }   \quad
        &
        \Pr \{ \bw^\T \bx_1 \geq b \} ,  \\ \notag
        \st \quad &
        \Pr \{ \bw^\T \bx_2 < b \} = 0.5.
    \end{align}
    They made the following two assumptions to solve \eqref{EQ:LAC_obj}:
    a) $\bw^\T \bx_1$ is Gaussian;
    b) $\bw^\T \bx_2$ is symmetric.
    By assuming these two assumptions, their objective function can be simplified to
    \begin{equation}
    \label{EQ:LAC_obj2}
          \max_{\bw} \frac{\bw^\T (\mu_1 - \mu_2)} { \sqrt{ \bw^\T \Sigma_1 \bw}  },
    \end{equation}
    where $\mu_1$ and $\mu_2$ are the mean of positive and negative classes, respectively.
    $\Sigma_1$ is the covariance matrix of positive classes.
    The form of \eqref{EQ:LAC_obj2} is similar to the
    LDA, which can be written as,
    \begin{equation}
    \label{EQ:LDA_obj2}
          \max_{\bw} \frac{\bw^\T (\mu_1 - \mu_2)}
                          { \sqrt{ \bw^\T (\Sigma_1 + \Sigma_2) \bw}  },
    \end{equation}
    where $\Sigma_2$ is the covariance matrix of negative classes.
    The only difference between LAC and LDA is that the pooled covariance matrix, $\Sigma_1$, is replaced by $\Sigma_1 + \Sigma_2$.
    \eqref{EQ:LAC_obj2} and \eqref{EQ:LDA_obj2} can be solved by eigen-decomposition and a closed-form solution for \eqref{EQ:LAC_obj2} and \eqref{EQ:LDA_obj2} can be derived as,
    \begin{equation}
    \label{EQ:LAC_opt_w}
        \bw^\ast_{\rm LAC} = \Sigma_1^{-1} (\mu_1 - \mu_2)
    \end{equation}
    and
    \begin{equation}
    \label{EQ:LDA_opt_w}
        \bw^\ast_{\rm LDA} = (\Sigma_1 + \Sigma_2)^{-1} (\mu_1 - \mu_2),
    \end{equation}
    respectively.
    It is important to note here that positive data, $\bx_1$, and negative data, $\bx_2$, are simply the output of weak classifiers.
    Hence, the solution expressed in \eqref{EQ:LAC_opt_w} can be used as a replacement for boosting coefficients and node rejection threshold.

    Nonetheless, LAC has several drawbacks.
    First, it relies on a limited set of features trained by AdaBoost.
    Shen \etal illustrate that when training data is highly asymmetric, AdaBoost can select a set of irrelevant features to re-initialize sample weights \cite{Shen2011Efficiently}.
    When this happens, LAC can only suppress weights of irrelevant features by setting their coefficients to be small.
    Second, LDA is shown to perform as well as LAC in object detection \cite{Shen2011Efficiently, Wu2008Fast}.
    In the next section, we present an efficient approach to prune
    irrelevant weak learners based on GSLDA/sparse eigenvectors.
    The algorithm integrates sparsity into the LDA classifier so that
    we only keep an optimal set of weak learners
    while achieving the asymmetric node learning goal in cascade classifiers.

    \subsection{An Efficient Pruning Algorithm Based on Sparse Eigenvectors}

    In the previous section, we discussed LAC and LDA which have empirically shown to be better at
    handling the asymmetric node learning goal in cascade classifiers.  In this section, we propose
    a more efficient feature search which finds a reduced set of hypotheses while satisfying
    \eqref{EQ:LAC_obj}.

    Pruning can be casted as a sparse representation problem.
    Sparse representation attempts to find a solution which uses only a small subset of original features.
    Sparse representation has been successfully applied to solve problems in many areas,
    \eg, signal compression \cite{Marcellin2000Overview},
    image de-noising \cite{Elad2006Image}, variable selection \cite{Tibshirani1996Regression},
    face recognition \cite{Wright2009Robust}, learning face features \cite{Destrero2009Sparsity},
    \etc.
    Motivated by the important role of sparsity,
    we prune the selected set of weak classifiers by training the sparse algorithm.
    Our objective is to set a coefficient vector, $\bw$, with many zero elements, indicating that only few of weak classifiers actually participate in the final decision rule.
    In this section, we propose to apply a sparse LDA to remove a set of irrelevant features.
    Similar to LDA, the sparse LDA solves the maximal class-separation problem but with an additional sparsity constraint.
    The objective function of sparse LDA can be written as,
    \begin{align}
    \label{EQ:GSLDA_obj}
        \max_{ \bw }   \quad
        &
        \frac{ \bw^\T S_b \bw } { \bw^\T S_w \bw }, \\ \notag
        \st \quad &
        \card(\bw) = k,
    \end{align}
    where $S_b$ and $S_w$ correspond to the between-class and within-class covariance matrices, respectively. $\card(\cdot)$ counts the number of non-zero components and $k$ is an integer set by the user.
    In our paper, we define the within-class covariance matrix as,
    \begin{align}
    \label{EQ:within_class_cov}
      S_w = \gamma \Sigma_1 + (1 - \gamma) \Sigma_2,
    \end{align}
    where $\Sigma_1$ and $\Sigma_2$ are covariance matrices of positive and negative classes, respectively.
    The parameter, $\gamma$, controls the weighted sum of covariance matrices between both classes.
    By setting $\gamma$ to $0.5$, we have the LDA objective \eqref{EQ:LDA_obj2} and by setting $\gamma$ to 1, we have the LAC objective \eqref{EQ:LAC_obj2}.
    In the experimental section, we conjecture that LDA is simply a regularized version of LAC.

    Due to the sparsity constraint in \eqref{EQ:GSLDA_obj}, a closed-form solution to LDA, \eqref{EQ:LDA_opt_w}, can no longer be used.
    \eqref{EQ:GSLDA_obj} can be solved using a branch-and-bound search to select a set of relevant features \cite{Moghaddam2006Generalized}.
    The algorithm finds an exact solution to the sparse problem.
    However, the algorithm is computationally expensive and is almost infeasible on large feature
    dimensions.  Even with a good initialization, the branch-and-bound search takes more than two
    hours to solve a problem where the size of the original feature space is $40$
    and the number of non-zero components, $k$, is set to $20$ \cite{Moghaddam2006Generalized}.
    In face detection, the number of Haar-like features
    can be more than $100,000$ and $k$ can be as large as $200$.
    Clearly, there is a need for a more efficient alternative solution.
    Two widely adopted approaches, to approximately solve the optimization problem with the sparsity constraint as in \eqref{EQ:GSLDA_obj},
    are forward and backward greedy algorithms.
    The algorithm sequentially selects a new feature/variable at each step to reduce the objective function.
    A forward selection has been commonly applied due to its effectiveness and efficiency.
    A shortcoming of forward selection is that the algorithm can never correct mistakes made in earlier steps.
    In order to remedy this situation, a backward greedy algorithm has been adopted.
    The idea is to train a full model and greedily remove one feature/variable at a time.
    In this paper, we adopt an efficient greedy approach proposed in \cite{Moghaddam2007Fast}.
    For our problem, the computation can be made very efficient as
    the objective of \eqref{EQ:GSLDA_obj} can be computed in a closed form as
    $\bb^\T  S_w^{-1}  \bb$
    due to the rank-$1$ $S_b$ matrix being a simple
    outer-product, $S_b =  \bb^\T \bb$.
    Therefore, the computational complexity is heavily determined by $S_w^{-1}$.
    A naive matrix inversion would be computationally expensive and inefficient.
    Since the matrix $S_w$ is sequentially appended or reduced by a single row and column,
    an efficient matrix inversion update algorithm can be exploitted \cite{Golub1996Matrix}.

    Let $A^t$ be a square symmetric matrix of size $t \times t$ and assume that we have computed its inverse,
    $(A^t)^{-1}$.
    If a vector, $\bv \in \mathbb{R}^{t+1}$, is appended to $A^t$ such that
    $A^{t+1} = \begin{bmatrix}
    A^t            & \bv_{(1:t)} \\
    \bv_{(1:t)}^\T & v_{(t+1)} \end{bmatrix}$.
    The new augmented inverse $(A^{t+1})^{-1}$
    can be calculated efficiently from
    \begin{align}
            \label{EQ:matrix_update_fwd}
            ( A^{t+1} ) ^ { -1 }  =
             \begin{bmatrix}
                                (A^t)^{-1}   +  a \bu \bu ^\T      &  - a \bu \\
                                - a \bu ^\T                        &  a
             \end{bmatrix},
    \end{align}
    where $ \bu = ( A^t )^{-1} \bv_{(1:t)}  $
    and  $ a = 1 /  (  v_{(t+1)} - \bv_{(1:t)} ^ \T \bu )$.
    Similarly, for backward greedy elimination, the new matrix inverse
    $(A^{t-1})^{-1}$ can be calculated by a simple rank-$1$ update as
    \begin{align}
            \label{EQ:matrix_update_backward}
            ( A^{t-1} ) ^ { -1 }  = B - (\bs_{(1:t-1)} \bs_{(1:t-1)}^\T) / s_{(t)},
    \end{align}
    where we partition the matrix inverse $( A^{t} ) ^ { -1 }$ as follows,
    $
            ( A^{t} ) ^ { -1 }  =
             \begin{bmatrix}
                                B               &  \bs_{(1:t-1)} \\
                                \bs_{(1:t-1)}^\T  &  s_{(t)}
             \end{bmatrix}
    $.
    Here we assume that we want to remove the last row and column of the matrix.
    Note that one would need to permute the row and column of the matrix if this
    is not the case.
    For a forward search, the greedy algorithm sequentially finds the suboptimal $\bw$ by adding a new variable which yields the
    maximal eigenvalue, $\bb^\T  S_w^{-1}  \bb$.
    On the other hand, for a backward elimination, the algorithm finds the suboptimal $\bw$
    by sequentially discarding a variable which yields the minimal eigenvalue.
    The algorithm continues until the predefined number of elements are selected,
    hence the name of greedy sparse LDA (GSLDA).  GSLDA is an excellent approach among other sparse
    algorithms due to its effectiveness and efficiency as shown in
    \cite{Moghaddam2006Generalized,Shen2011Efficiently}.

    \begin{figure}[t]
    \centering
        \includegraphics[width=0.4\textwidth,clip]{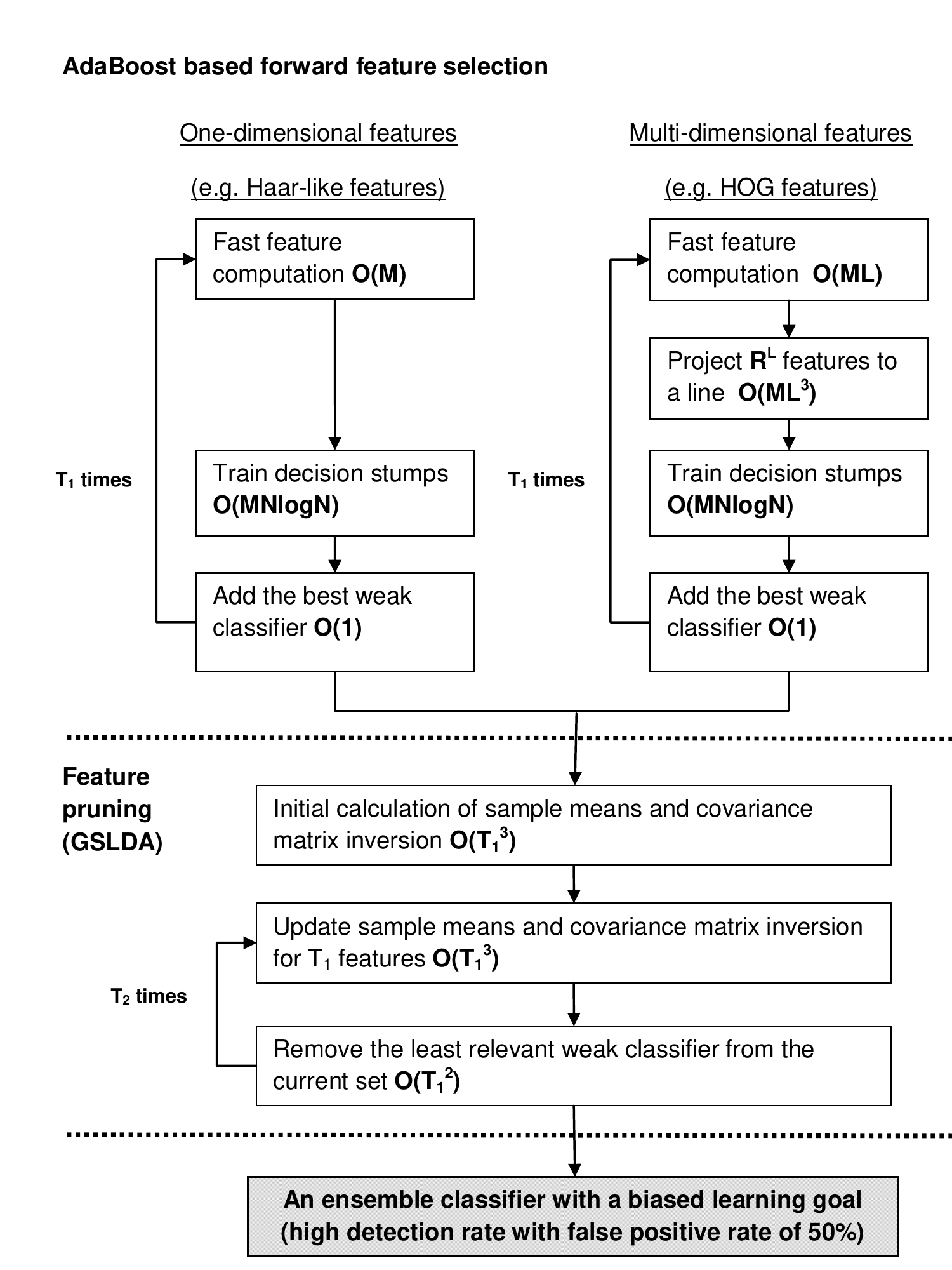}
    \caption{Flowchart of the proposed pruning algorithm.
    }
    \label{fig:flowchart}
    \end{figure}

    We illustrate the flowchart of our approach in Fig.~\ref{fig:flowchart}.
    Our algorithm works as follows.
    We first train a pool of discriminative features using AdaBoost.
    We then prune selected weak classifiers by performing a sequential backward elimination.
    The set of features which less satisfies the objective criterion \eqref{EQ:LDA_obj2}
    will be removed from our feature sets.
    Backward elimination continues until the required ensemble size is reached or the predefined node learning goal is achieved, \eg, $99\%$ detection rate and $50\%$ false positive rate.
    Finally, we adjust the threshold of the node classifier such that they have
    $50\%$ false positive rate on the training set.
    The final classifier will have a similar form as AdaBoost \eqref{EQ:FxAda}.
    We summarize the algorithm of our pruning approach in Algorithm~\ref{ALG:pruning}.

\SetKwInput{KwInit}{Initilaize}

\SetVline
\linesnumbered

\begin{algorithm}[t]
\caption{The pruning algorithm.}
\begin{algorithmic}
\footnotesize{
   \KwIn{
   \begin{itemize}
      \item
         A set of examples $\{\bx_i,y_i\}$, $i=1 \cdots N$;
      \item
         The number of initial weak classifiers to be trained, $T_1$;
      \item
         The maximum number of weak classifiers for the given node, $T$;
   \end{itemize}
   }

   \KwOut{
\begin{itemize}
   \item
      An ensemble classifier, $F(\bx) = \sign \bigl( \sum_{j=1}^{T} \alpha_j h_j(\bx) - b \bigr)$,
      that best satisfies the asymmetric learning objective \eqref{EQ:LAC_obj};
\end{itemize}
}

\KwInit {

   \begin{itemize}
      \item
         $t \leftarrow 0$;
      \item
         Initialize sample weights;
   \end{itemize}
}

\While{ $t < T_1$ (Selecting weak classifiers using AdaBoost)}         
{
  1. Train a weak learner (\eg,
  a decision stump, \eqref{EQ:2}, %
  that results in the smallest misclassification error)
  on the training set;
  \\2. Add the best weak learner into the current set;
  \\3. Update sample weights based on AdaBoost;
  \\4. $t \leftarrow t + 1;$
}

\While{ $t > T$ (Pruning using GSLDA)}
{
  1. Remove the weak classifier that least satisfies the asymmetric node learning goal
  \eqref{EQ:LAC_obj},
  using the GSLDA algorithm;
  \\2. $t \leftarrow t - 1;$
}

Adjust the threshold value $b$ such that $F$ has a $50\%$
   false positive rate on the training set.

} %
\end{algorithmic}
\label{ALG:pruning}
\end{algorithm}

    Note that \cite{Shen2011Efficiently} also used GSLDA to perform feature selection on a set of Haar-like features.
    However, the set of stump features used in their work is much smaller compared to ours. In their work, the number of stump features is equal to the number of Haar-like features, \ie, stump features are generated by training a decision stump on each Haar-like feature with uniform sample weights.
    In our approach, AdaBoost searches entire stump spaces and keeps a set of potential stump features.
    Hence, the size of our feature space is much larger than the one they adopted.
    Compared to \cite{Shen2011Efficiently}, our approach is not only more discriminative but also faster to train.
    In our experiments, we train $200 \sim 500$ weak classifiers using AdaBoost.
    These selected weak classifiers ensure an over-complete and optimal set of candidates
    for the asymmetric node learning goal.

\subsection{Time and Memory Complexity}

\begin{table}[tb!]
  \caption{A computational and memory complexity for our pruning algorithm.}
  \centering
  \scalebox{0.736}{
    \begin{tabular}{l|c|c}
    \hline
      $ $ & \textbf{Time}& \textbf{Memory} \\
    \hline
    \hline
      Feature acquisition (Haar-like) & $\bigO(T_1 M)$ & $\bigO(ND)$ \\
      Feature acquisition (HOG) & $\bigO(T_1 M L^3)$ & $\bigO(NLD)$ \\
      AdaBoost & $\bigO(T_1 M N \log N)$ & $\bigO(N)$ \\
      Pruning & $\bigO(T_2 T_1^3)$ & $\bigO(T_1^2)$ \\
      \hline
      \textbf{Total} (Haar-like features) & $\bigO(T_1 M+T_1 M N \log N + T_2 T_1^3)$ & $\bigO( ND + N + T_1^2)$ \\
      \textbf{Total} (HOG features) & $\bigO(T_1 M L^3 + T_1 M N \log N+ T_2 T_1^3)$ & $\bigO(NLD+ N + T_1^2)$ \\
     \hline
    \end{tabular}
    }
  \label{tab:complexity}
\end{table}

    To analyze the complexity, we break our approach into three stages: feature acquisition,
    AdaBoost and pruning.
    We first analyze the complexity of acquiring low level features, \eg, Haar-like features
    \cite{Viola2004Robust} or histogram of oriented gradients (HOG) features
    \cite{Dalal2005Histograms}. One may also use features like covariance features
    \cite{Paul2008Cov,Tuzel2008Cov}  or CENTRIST \cite{Wu2011}.
    In this step, we pick features, which can be computed in linear time with the use of integral
    images \cite{Wu2008Fast} or integral histograms \cite{Porikli2005Integral}.
    Hence, this step costs $\bigO(M)$ time for one-dimensional features and $\bigO(ML)$ time for multi-dimensional features.
    For multi-dimensional features, we project computed features onto a line using Fisher Linear Discriminant Analysis (LDA).
    LDA can be efficiently solved by generalized eigenvalue decomposition.
    This additional step costs $\bigO(ML^3)$ where $L$ is the number of dimensions (the total number of histogram bins for a block of HOGs).
    Hence, the feature computation step takes $\bigO(M)$ time for one-dimensional features and
    $\bigO(ML^3)$ time for multi-dimensional features (since $\bigO(ML + ML^3) \in \bigO(ML^3)$).
    For memory complexity, we need to store integral images for each training sample.
    For one-dimensional features, each training sample has a memory complexity of $\bigO(D)$ and the total memory complexity for $N$ training samples is $\bigO(ND)$.
    For multi-dimensional features, each training sample has a memory complexity of $\bigO(LD)$ and the total memory complexity is $\bigO(NLD)$.

    In the next step, we train weak classifiers known as decision stumps, \eqref{EQ:2}.
    To train decision stumps, we have to find the optimal threshold, $\theta$ in \eqref{EQ:2}, which produces a minimal misclassification error.
    For fast training of decision stumps, one can sort feature values and scan through all possible threshold values sequentially to update error rate of decision stumps \cite{Wu2008Fast}.
    This algorithm takes $\bigO(N \log N)$ for sorting and $\bigO(N)$ for scanning.
    We ignore $\bigO(N)$ since $\bigO(N \log N)$ is bigger than $\bigO(N)$.
    In each AdaBoost iteration, we need to train $M$ decision stumps.
    Hence, this step takes $\bigO(M N \log N)$.
    Training $T_1$ iterations of AdaBoost would take $\bigO(T_1 M N \log N)$.
    For memory complexity, we need to store sorted feature values, $\bigO(N)$, and the output ensemble classifiers, $\bigO(T_1)$.
    Hence, the total complexity is $\bigO(N+T_1)$ memory.
    In object detection, we often have $N \gg T_1$, which means AdaBoost requires approximately $\bigO(N)$ memory.
    In summary, the entire AdaBoost training has a complexity of $\bigO(T_1 M N \log N)$ time and $\bigO(N)$ memory.

    In the final step, we prune weak classifiers obtained from AdaBoost using backward elimination.
    Backward elimination starts with the full index set ($T_1$) and sequentially deletes the variable which is the least relevant until only $T$ elements remain.
    To begin a pruning operation, we first compute sample means and covariances for both positive and negative samples.
    We then calculate the inverse of covariance matrices.
    This process requires $\bigO(T_1^3)$.
    Next, we sequentially remove less discriminative features from the current set.
    Each backward elimination step costs $\bigO(T_1^3)$ for matrix inversion update and $\bigO(T_1^2)$ for removing the least relevant feature from the current set.
    Since we perform this step $T_2$
    times\footnote{$T_2$ is the number of features to be removed during
    pruning, \ie, $T_2 = T_1 - T$, where
    $T_1$ is the number of initial features to be selected and
    $T$ is the number of features in the node classifier.},
    the total computational cost of backward search is $\bigO(T_2T_1^2 + T_2 T_1^3)$, which is
    $\bigO(T_2 T_1^3)$ time.
    For memory complexity, we need to store both sample means, $\bigO(T_1)$, and sample covariances, $\bigO(T_1^2)$. The total memory complexity is $O(T_1^2)$.
    Hence, the computation cost of pruning has a total complexity of $\bigO(T_2 T_1^3)$ time and $\bigO(T_1^2)$ memory.
    It is important to point out here that $M \gg T_1$ and $N \gg T_1$ in vision tasks.
    By not having $M$ or $N$ in the computational complexity of pruning,
    pruning does not take up majority of the training time.
    In the experimental section, we show that most of the training time is actually spent
    on bootstrapping difficult negative samples in latter cascade nodes.
    Table~\ref{tab:complexity} summarizes the complexity in terms of time and memory of our approach.

\section{Experiments}
\label{sec:exp}

    \subsection{Face Detection on a Single-node Detector}

    \begin{figure}[t]
    \centering
        \includegraphics[width=0.35\textwidth,clip]{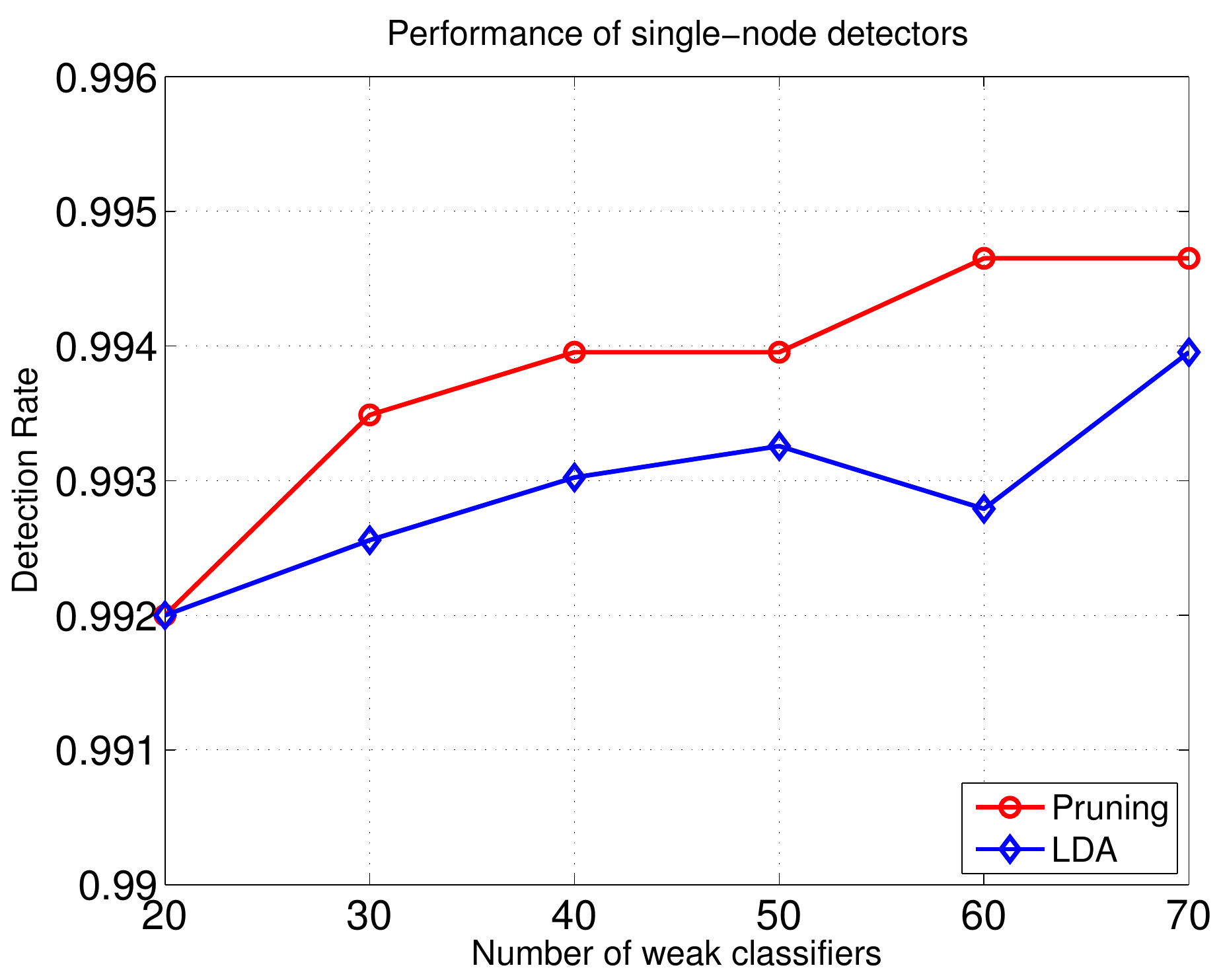}
        \\
        \includegraphics[width=0.35\textwidth,clip]{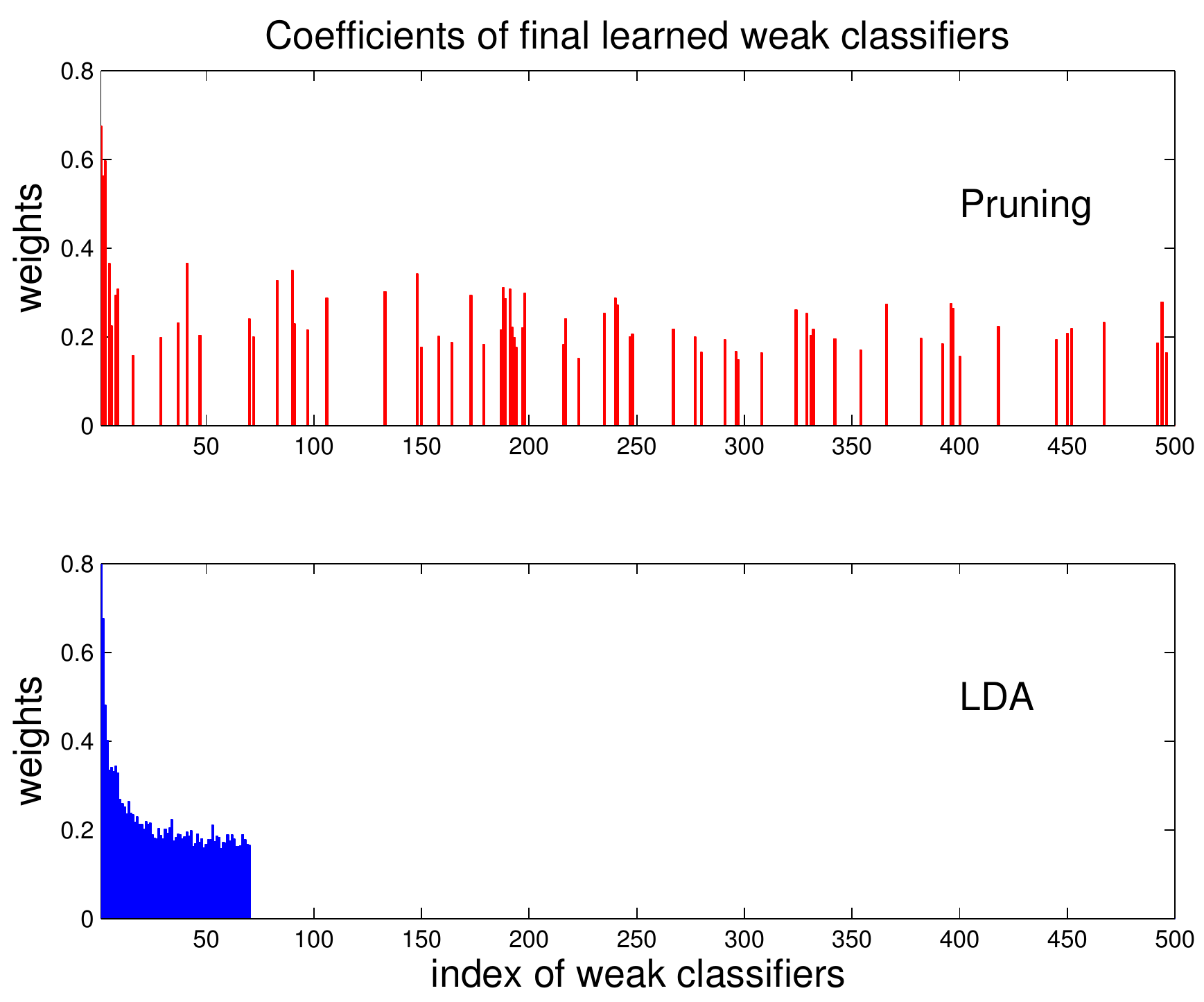}
    \caption{Detection performance (top) of a single-node classifier.
    Here ``Pruning'' is our approach and ``LDA'' is AdaBoost+LDA approach
    \cite{Wu2008Fast}.
    We measure the detection rate by setting the false positive rate
    on the test set to $50\%$.
    Our approach has a higher detection rate than AdaBoost+LDA of
    Wu \etal.
    The bottom plot shows values of weak classifiers' coefficients for both
    methods. Our pruning approach selects very different weak classifiers compared
    with Wu \etal.
    }
    \label{fig:singlenode}
  \end{figure}

  \begin{figure}[t]
    \begin{center}
        \includegraphics[width=0.47\textwidth,clip]{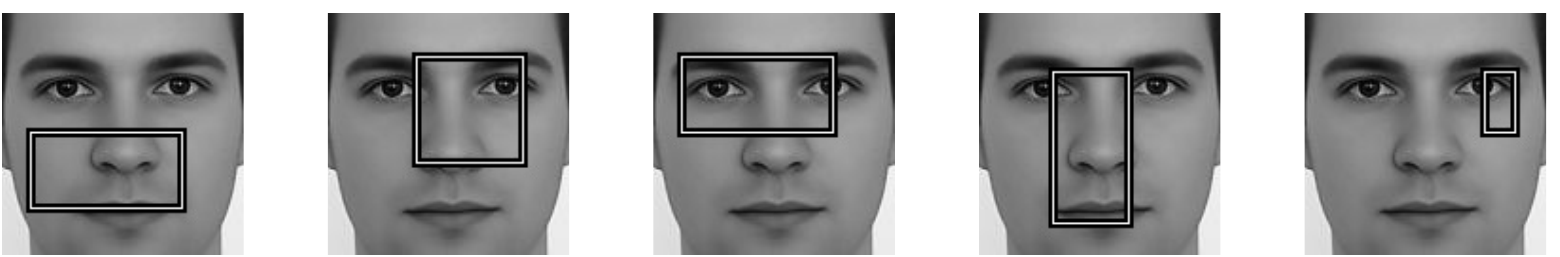}
        \\
        \includegraphics[width=0.47\textwidth,clip]{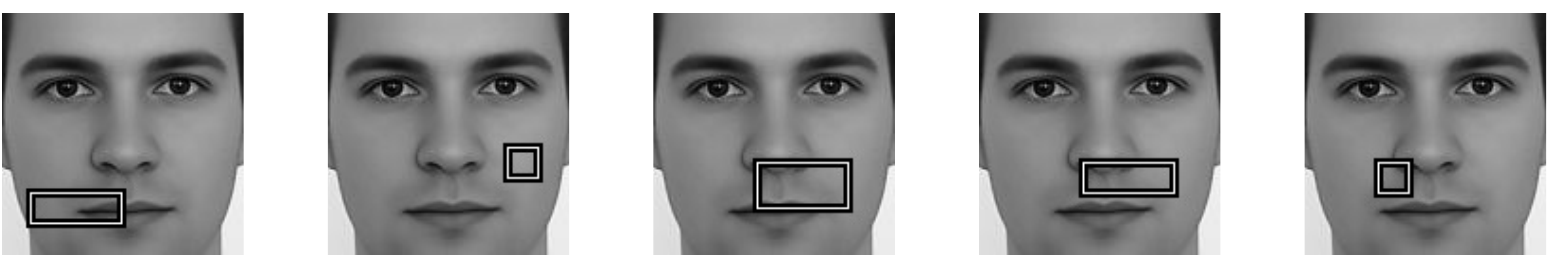}
        \\
        \includegraphics[width=0.47\textwidth,clip]{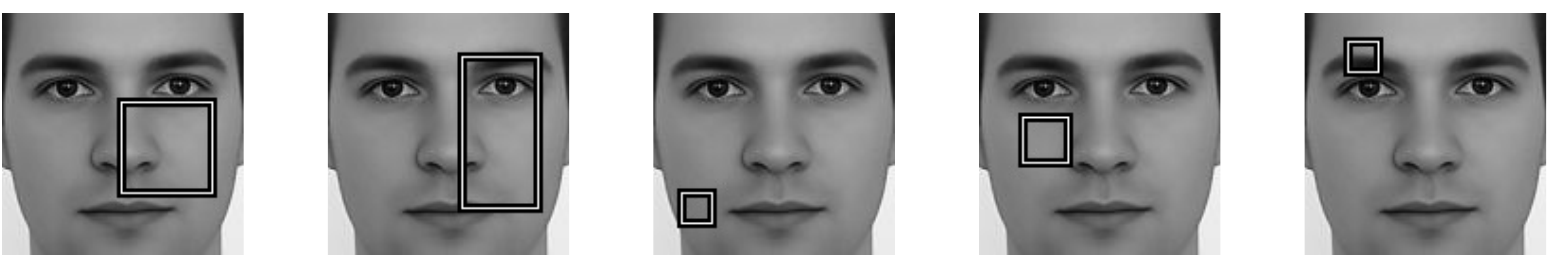}
    \end{center}
    \caption{Illustration of selected and discarded HOG blocks
    using the asymmetric node learning criterion.
    {Top:} Kept HOG blocks;
    {Middle:} Removed HOG blocks by pruning from the first 70 weak
    learners selected by AdaBoost;
    {Bottom:} HOG blocks that are kept by our pruning approach but not
    in the first 70 weak learners selected by AdaBoost.
    }
    \label{fig:singlenode_rm}
  \end{figure}

    The aim of our first experiment is to emphasize the performance difference between AdaBoost+LDA
    \cite{Wu2008Fast} and our pruning approach.
    In this experiment, we set the value of $\gamma$ to
    $0.5$ (LDA).
    We use the FDDB face detection benchmark data set \cite{Jain2010FDDB}.
    The data set consists of $5,171$ faces in $2,845$ images under varying conditions in unconstrained environments
    (faces in the wild).
    Here we use  HOG   features to train the face detector.
    We discard face samples which have a resolution less than $48 \times 48$ pixels.
    The rest of faces is scaled to $48 \times 48$ pixels with $8$ pixels additional border to
    preserve the contour information of faces.
    $9,300$ remaining faces are split into two sets.
    The first set contains $5,000$ faces and $5,000$ random non-faces for training.
    The second set contains $4,300$ faces and $100,000$ random non-faces for evaluation.
    In our experiment, we use blocks with different scales ($8 \times 8$ pixels to $48 \times 48$
    pixels) and various aspect ratios ($1:1$, $1:2$, $2:1$, $1:3$ and $3:1$).
    Each block is divided into $2 \times 2$ cells and the HOG in each cell is summarized into $9$
    bins \cite{Dalal2005Histograms}.
    Hence, $36$-dimensional features are generated for each block.
    An $\ell_1$-Sqrt normalization is applied to the feature vector.
    At each iteration, we randomly sample $25\%$ of all possible blocks for training a weak
    classifier.
    For our approach, we use AdaBoost to select 500 weak classifiers in the first
    step, which is assumed to contain most discriminative weak classifiers for this task.
    Fig.~\ref{fig:singlenode} shows the detection rate of both algorithms by fixing the false
    positive rate to $50\%$.  In other words, each algorithms are programmed to remove $50,000$
    non-faces.
    From the figure, our approach has a higher detection rate performance
    than AdaBoost+LDA \cite{Wu2008Fast}.
    Our algorithm also uses a smaller number of weak classifiers.
    For example, to achieve a $99.4\%$ detection rate and $50\%$ false positive rate on test sets,
    our algorithm requires $40$ weak classifiers (pruned from $500$ weak classifies)
    while AdaBoost+LDA needs at least $70$ weak classifiers.
    Fig.~\ref{fig:singlenode} also shows the value of weak classifiers' coefficients,
    $\bw$, of both
    algorithms.  The $x$-axis represents the index of all $500$ candidate weak classifiers.  The
    $y$-axis represents weights of weak classifiers.  It can be observed that our approach
    selects very different weak classifiers compared with AdaBoost+LDA.

    We also illustrate HOG blocks that are selected and discarded
    using the proposed approach.
    Fig.~\ref{fig:singlenode_rm} (top) shows five HOG blocks with
    the highest weak learners' coefficients which are kept by
    our approach (in the first 70 weak learners selected by AdaBoost).
    The middle row shows five less relevant HOG blocks (with the lowest weak
    learners' coefficients) which are removed by pruning from the set of
    first 70 selected weak learners by AdaBoost.
    The bottom row shows five HOG blocks kept by the pruning approach from the pool of 500 weak
    learners but not in the first 70 selected by AdaBoost.

    We observe that blocks which cover the lower part of the face,
    \ie, areas around cheeks, are often removed.
    On the other hand, HOG blocks located around the eyes and nose are more likely
    to be kept or selected by our pruning approach.
    Clearly, our pruning approach
    has a higher flexibility in choosing a set of discriminative weak classifiers than AdaBoost+LDA.
    As demonstrated in this experiment,
    the first $70$ weak classifiers selected by AdaBoost is not necessarily the optimal set
    that meets the asymmetric node learning goal.
    Consequently, it also justifies the need of using AdaBoost to select a large over-complete set
    of weak classifiers.

   \subsection{Face Detection on MIT-CMU}

   \begin{figure}[t]
    \centering
        \includegraphics[width=0.35\textwidth,clip]{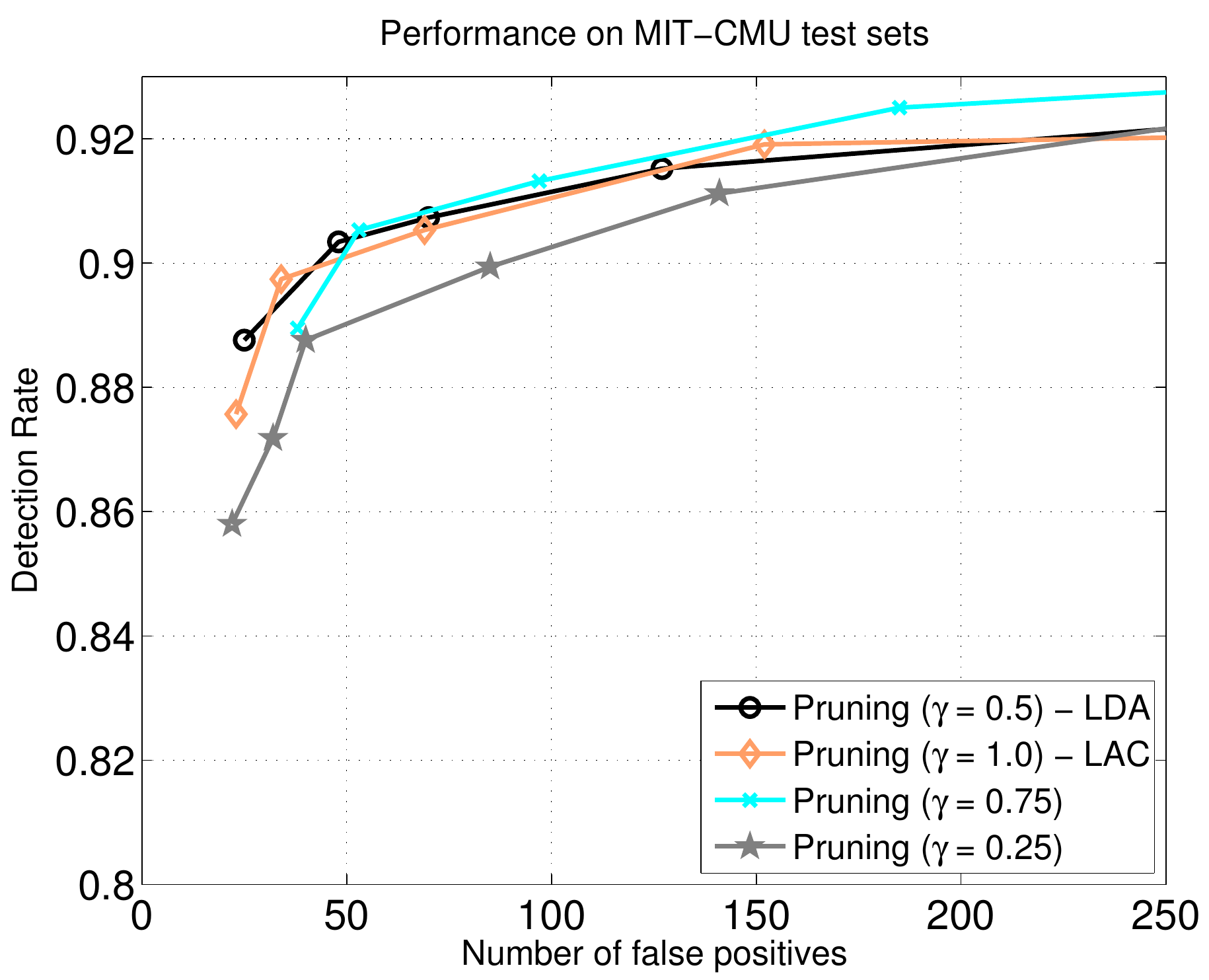}
        \includegraphics[width=0.35\textwidth,clip]{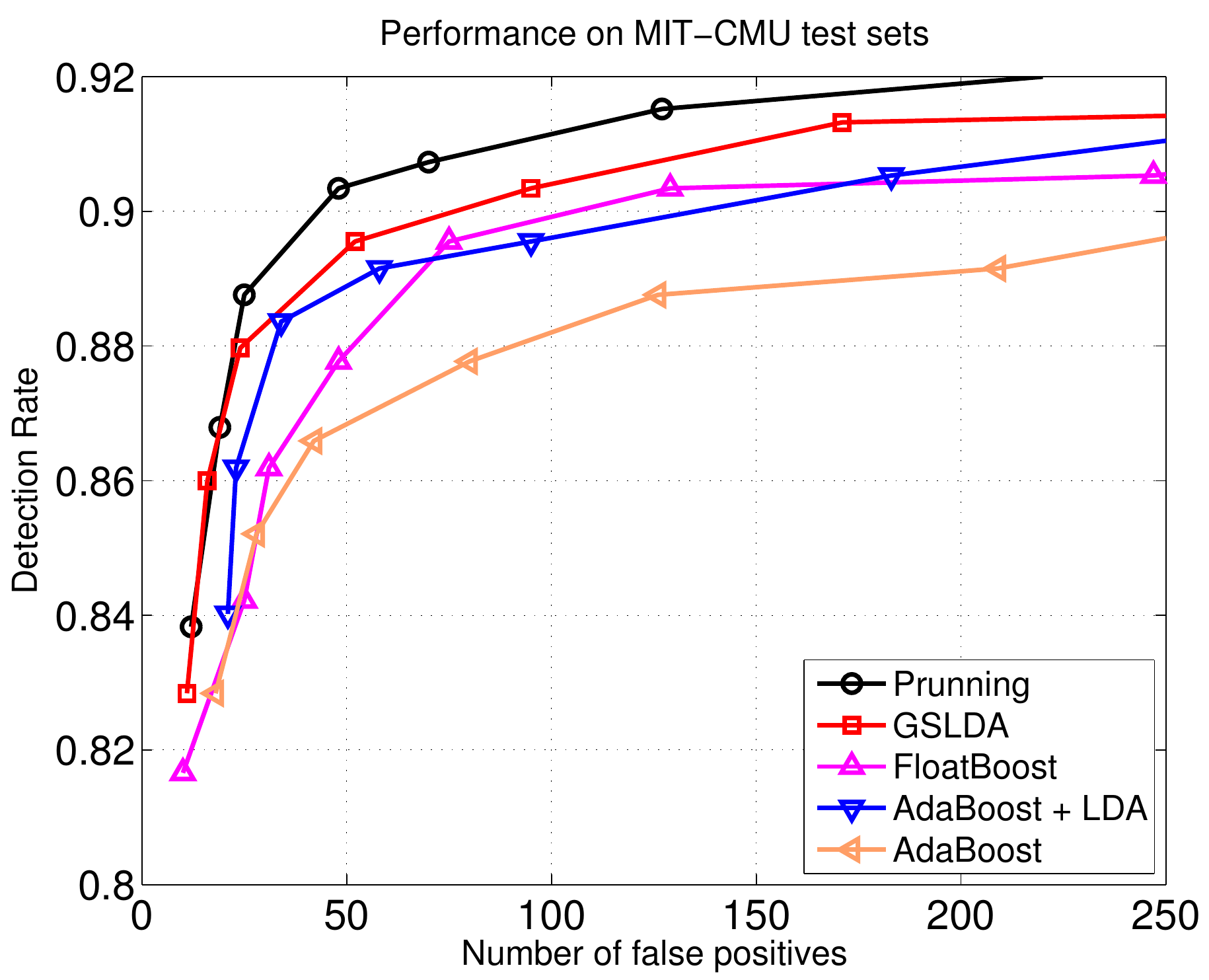}
    \caption{Performance comparison on MIT+CMU face data sets.
             We evaluate our approach with different pruning
             parameters (top) and
             compare against several face detectors (bottom).
    }
    \label{fig:mitcmu}
  \end{figure}

  \begin{figure*}[t]
    \begin{center}
        \includegraphics[width=0.65\textwidth,clip]{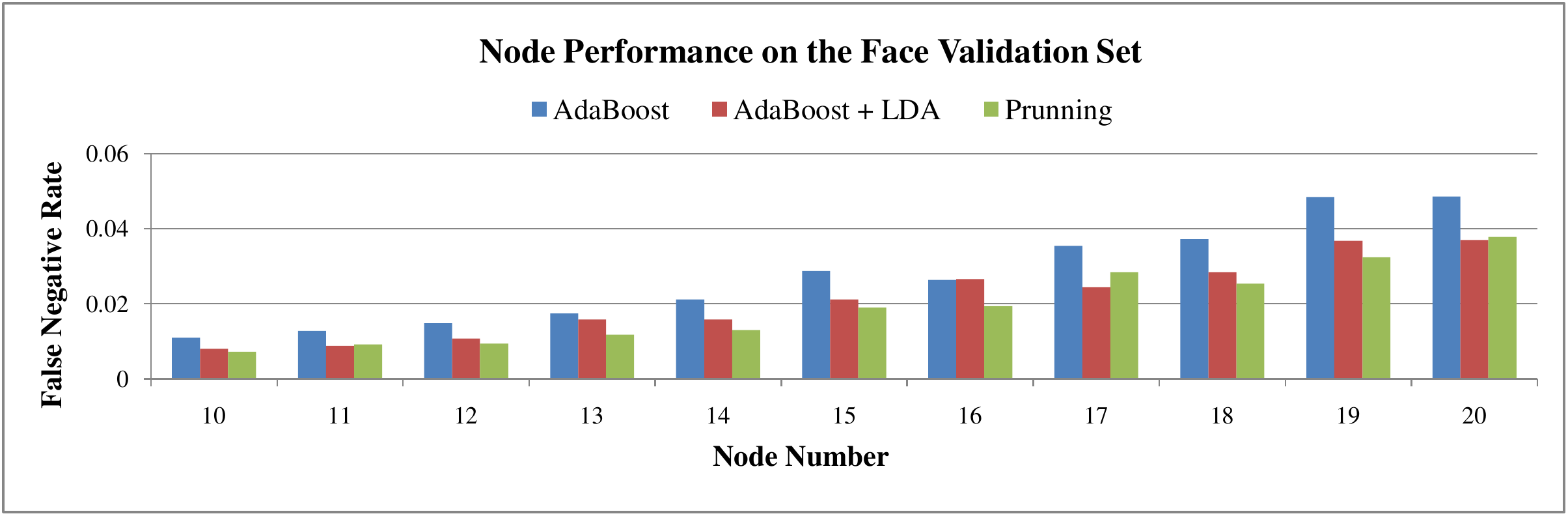}
    \end{center}
    \caption{Independent node performance on validation sets for face detection.
    In most nodes, our approach results in a smaller false negative rate.
    }
    \label{fig:mitcmu_node}
  \end{figure*}

    In this experiment, we first
    evaluate our approach on face detection with different values of $\gamma$
    \eqref{EQ:within_class_cov}.
    In this experiment, we set $\gamma \in \{0.25, 0.5, 0.75, 1.0\}$.
    Our training set contained $5,000$ face patches and
    $5,000$ initial non-face patches.
    The resolution of the training data is $24 \times 24$ pixels.
    Negative patches are collected from $10,000$ background images.
    We used $16,233$ features sampled uniformly from the entire set of Haar-like features.
    We train $20$ node classifiers.
    The number of weak classifiers in each node is $7$, $15$, $30$, $30$,
    $50$, $50$, $50$, $100$, $120$, $140$, $160$, $180$, $200$, $200$, $\cdots$, $200$.
    In all nodes, we adjust the threshold such that each node achieves $50\%$
    false positive rate on the training data.
    For our pruning approach, we set $T_1$ to $500$.
    In other words, we collect bootstrapped non-face patches
    and train $500$ weak classifiers using AdaBoost in each node.
    Note here that it is possible to set $T_1$ to be larger.
    Doing so would guarantee that the initial set of features is over-complete.
    However, this also increases the overall computational time during training.
    Our cascade training algorithm terminates when the
    bootstrapping non-face image database is depleted.
    During evaluation,
    the test image is re-scaled repeatedly by a factor of $1.25$ and scanned with a stride of 1 pixel in both directions.
    Post-processing similar to \cite{Viola2004Robust} is applied to merge final detection results.
    We construct receiver operating characteristic (ROC) curves by repeatedly removing nodes from
    the cascade to generate points with increasing detection and false positive rates.  Results are
    shown in Fig.~\ref{fig:mitcmu}.
    Based on ROC curves, $\gamma \in \left[0.5, 1.0\right]$ perform similarly.
    We suspect that the LDA criterion ($\gamma = 0.5$)
    performs similarly to the LAC criterion ($\gamma = 1$) since
    LDA is simply a regularized version of LAC.
    In other words, if we assume that the covariance matrix
    of the negative data is approximately $\nu \I$, \ie, $\Sigma_2 \approx \nu \I$, where $\nu$
    is a positive constant and $\I$ is an identity matrix.
    \eqref{EQ:within_class_cov} can be written as,
    \begin{align}
    \label{EQ:within_class_cov2}
      S_w = \Sigma_1 + \frac{1 - \gamma}{\gamma} \Sigma_2 = \Sigma_1 + \nu \I.
    \end{align}
    In our experiments, we observe this assumption to be valid for the
    asymmetric node learning objective, \ie,
    the off-diagonal elements of $\Sigma_2$ (correlation values) is often
    close to zero.
    Our conjecture is that negative data in each node are bootstrapped
    from a large pool of background images and likely to follow
    a uniform distribution.  Hence, LDA is simply a
    regularized LAC.  It can be argued that the distribution of negative data may not follow the
    uniform distribution in latter stages where a large number of background patches have already
    been filtered out.  In this case, we conjecture that it may be best to use covariance information from both positive
    and negative data ($\gamma < 1$).  This might explain why Wu \etal also observed that, in some
    cases, LDA gives a better performance than LAC \cite{Wu2008Fast}.  In the rest of our
    experiments, we set $\gamma$ to $0.5$ (LDA).

    In the next experiment, we compare our approach with AdaBoost
    \cite{Viola2004Robust}, AdaBoost with the LDA
    post-processing (AdaBoost+LDA) \cite{Wu2008Fast},
    AdaBoost with a backward elimination
    (FloatBoost) \cite{Li2004Float} and the GSLDA algorithm
    \cite{Shen2011Efficiently}.  We compare the
    performance of five object detectors on the MIT+CMU test sets.
    All experimental settings remain the same as described previously, \ie,
    all five cascade classifiers are trained with the same number of
    haar-like features and the node learning goal
    of all five detectors are set to be the same.
    ROC curves are plotted in Fig.~\ref{fig:mitcmu}.
    Note here that the experimental result of other detectors
    are based on our own re-implementation.
    Experimental results demonstrate that pruning AdaBoost with the LDA criterion performs best.
    It outperforms the GSLDA object detector as it incorporates more relevant features.
    Our proposed classifier also outperforms FloatBoost
    as FloatBoost does not consider the asymmetry in the learning.

    Fig.~\ref{fig:mitcmu_node} shows an independent node comparison on $4,832$ test faces with different
    learning algorithms.
    We evaluate each node independently.
    From the figure, pruning has a smaller false negative rate (higher detection rate) than AdaBoost
    and AdaBoost+LDA in most nodes.
    In summary, our results clearly demonstrate the superior performance of pruning.
    Table~\ref{tab:traintime} shows the approximate cascade training time and the average number of
    features evaluated per patch.
    Note that with the use of integral images and caching, training
    each node of cascade classifiers takes less than 5 minutes.
    We observe that most of the training time
    is spent on bootstrapping difficult negative samples in latter cascade nodes.
    Our experiments are performed on a server with $12$-core
    AMD Opteron of $2.20$ GHz and $256$ GB RAM.
    The code is implemented in C++ and OpenMP API for parallelized feature extraction,
    feature sorting and bootstrapping.

\begin{table}[t!]
  \caption{Comparison of training time and the average number of features evaluated per patch for different detectors.
  The average number of features used was obtained from test sets indicated.
   }
  \begin{center}
      \scalebox{.95}{
    \begin{tabular}{l|c|c|c|c}
    \hline
      & \multicolumn{2}{|c|}{ MIT-CMU (Haar)} &  \multicolumn{2}{|c}{ FDDB (HOG) } \\
    \cline{2-5}
    $ $ & Train. & $\#$ Evaluation & Train. & $\#$ Evaluation \\
    \hline
    \hline
      Pruning                & $6$h$10$m & $16.31$ & $4$h$30$m & $15.07$ \\
      AdaBoost \cite{Viola2004Robust} & $5$h$15$m & $13.75$ & $3$h$30$m & $13.89$  \\
      AdaBoost+LDA \cite{Wu2008Fast}  & $5$h$25$m & $13.09$ & $3$h$40$m & $13.88$  \\
     \hline
    \end{tabular}
    }
\end{center}
  \label{tab:traintime}
\end{table}

\subsection{Challenging Face Detection Data Sets}

  \begin{figure}[t]
    \begin{center}
        \includegraphics[width=0.17\textwidth,clip]{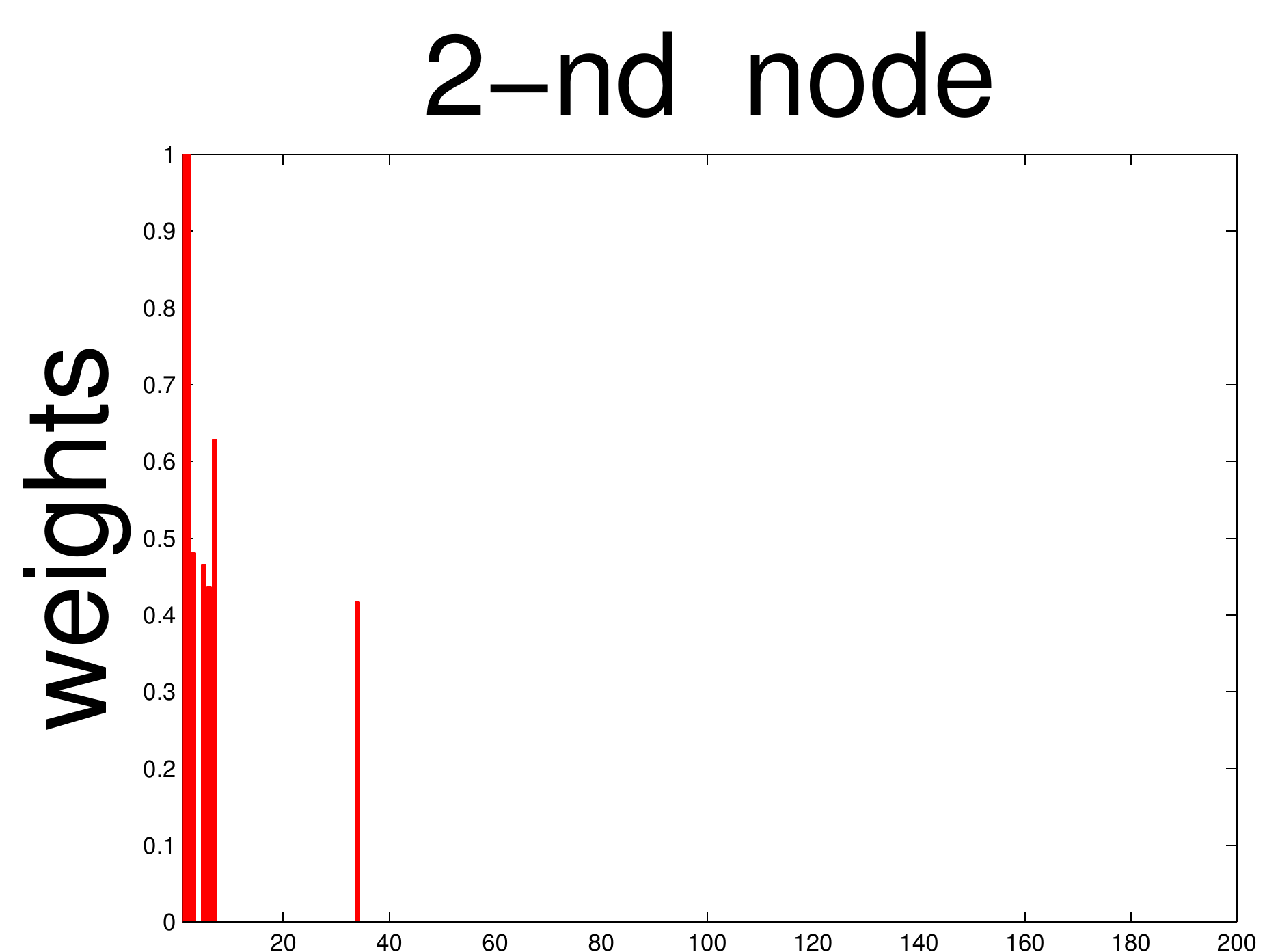}
        \includegraphics[width=0.15\textwidth,clip]{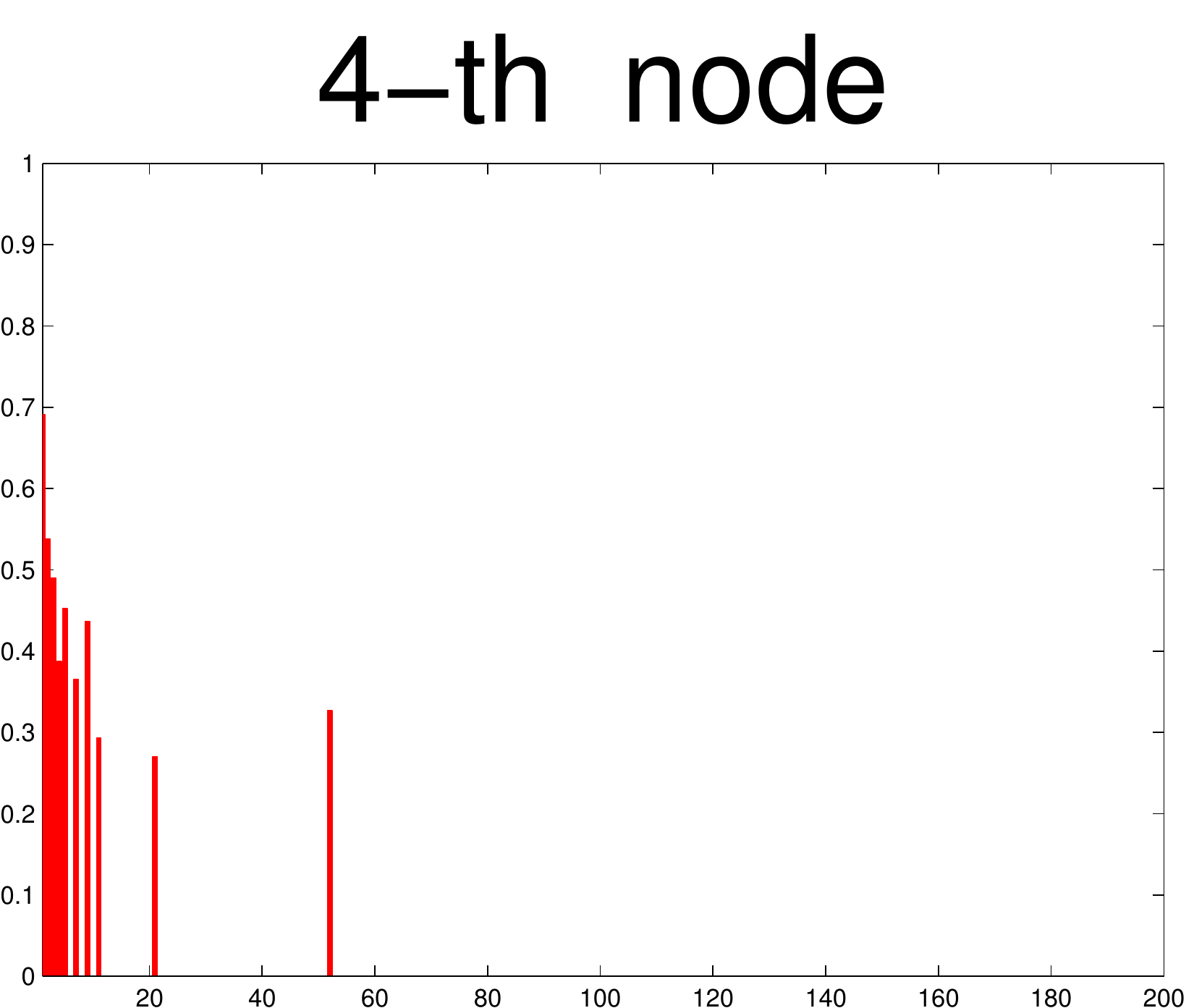}
        \includegraphics[width=0.15\textwidth,clip]{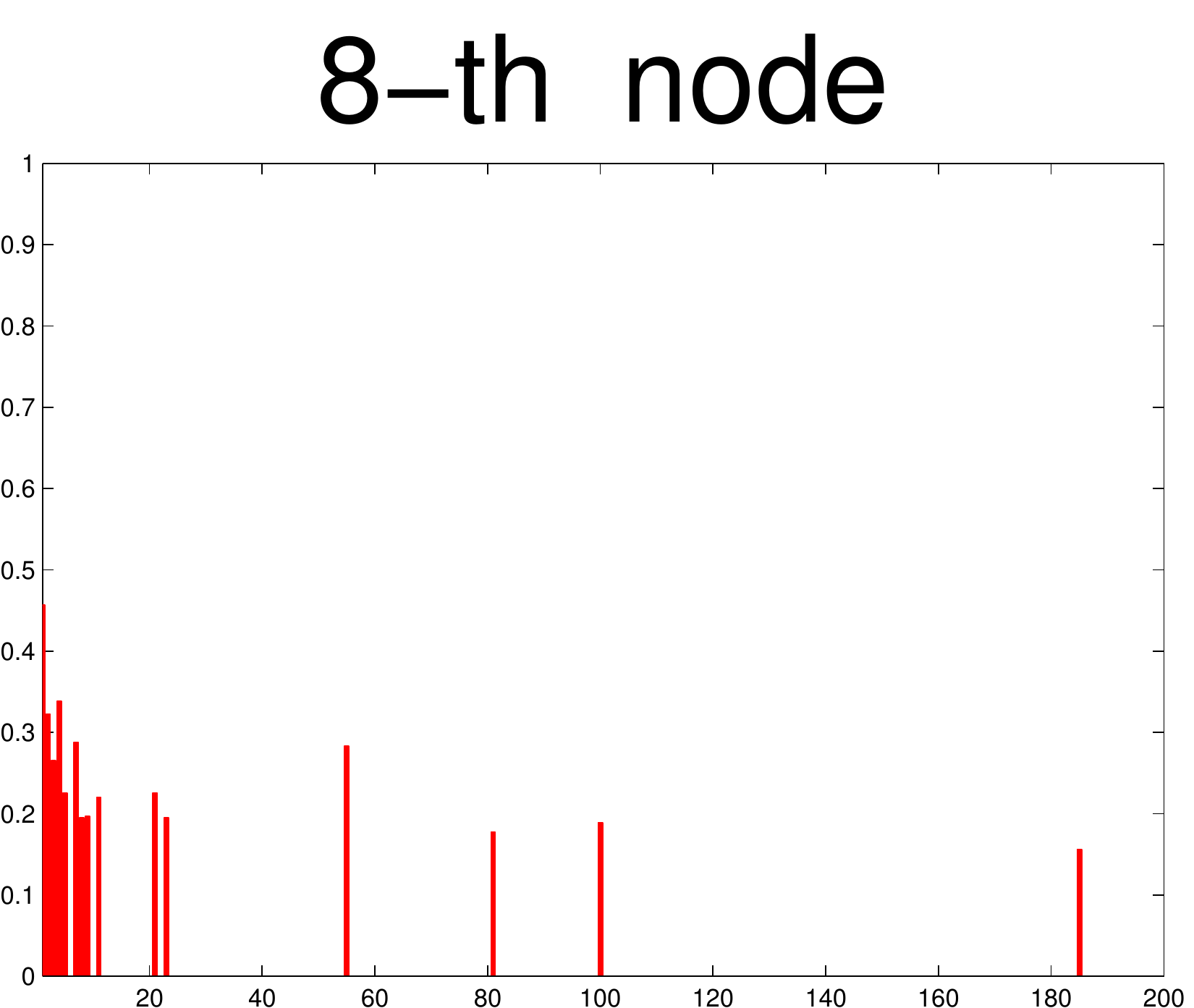}
        \includegraphics[width=0.17\textwidth,clip]{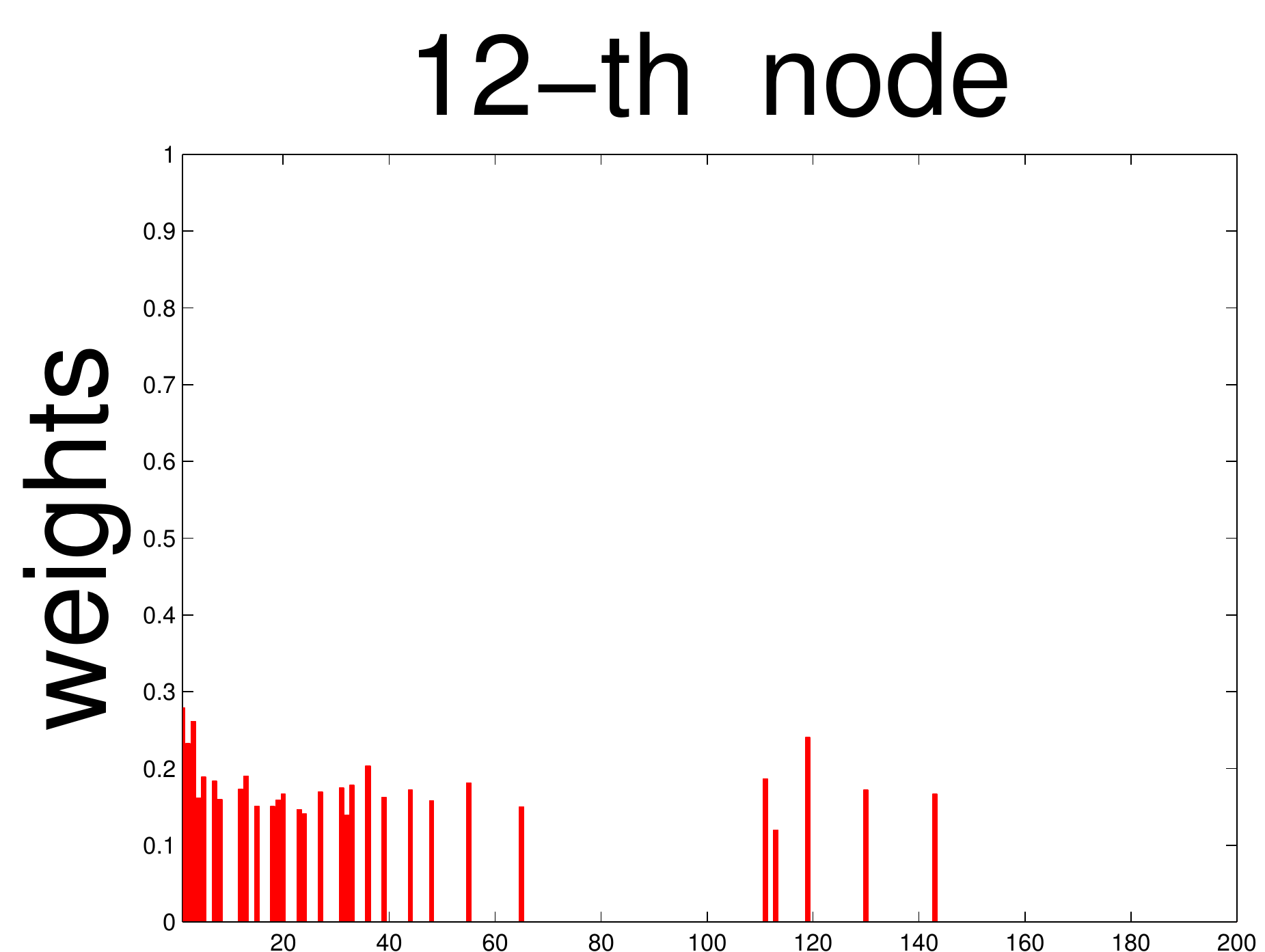}
        \includegraphics[width=0.15\textwidth,clip]{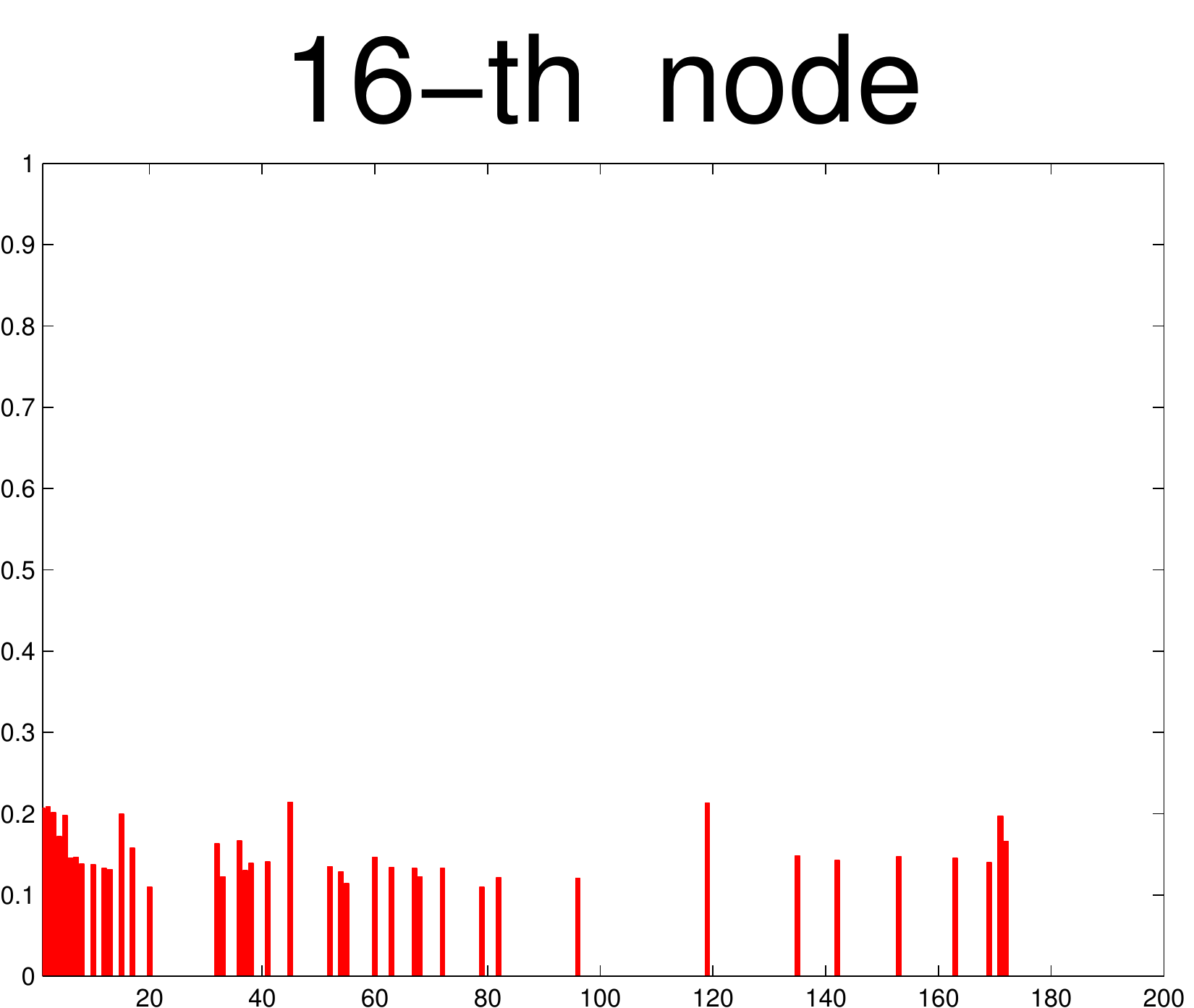}
        \includegraphics[width=0.15\textwidth,clip]{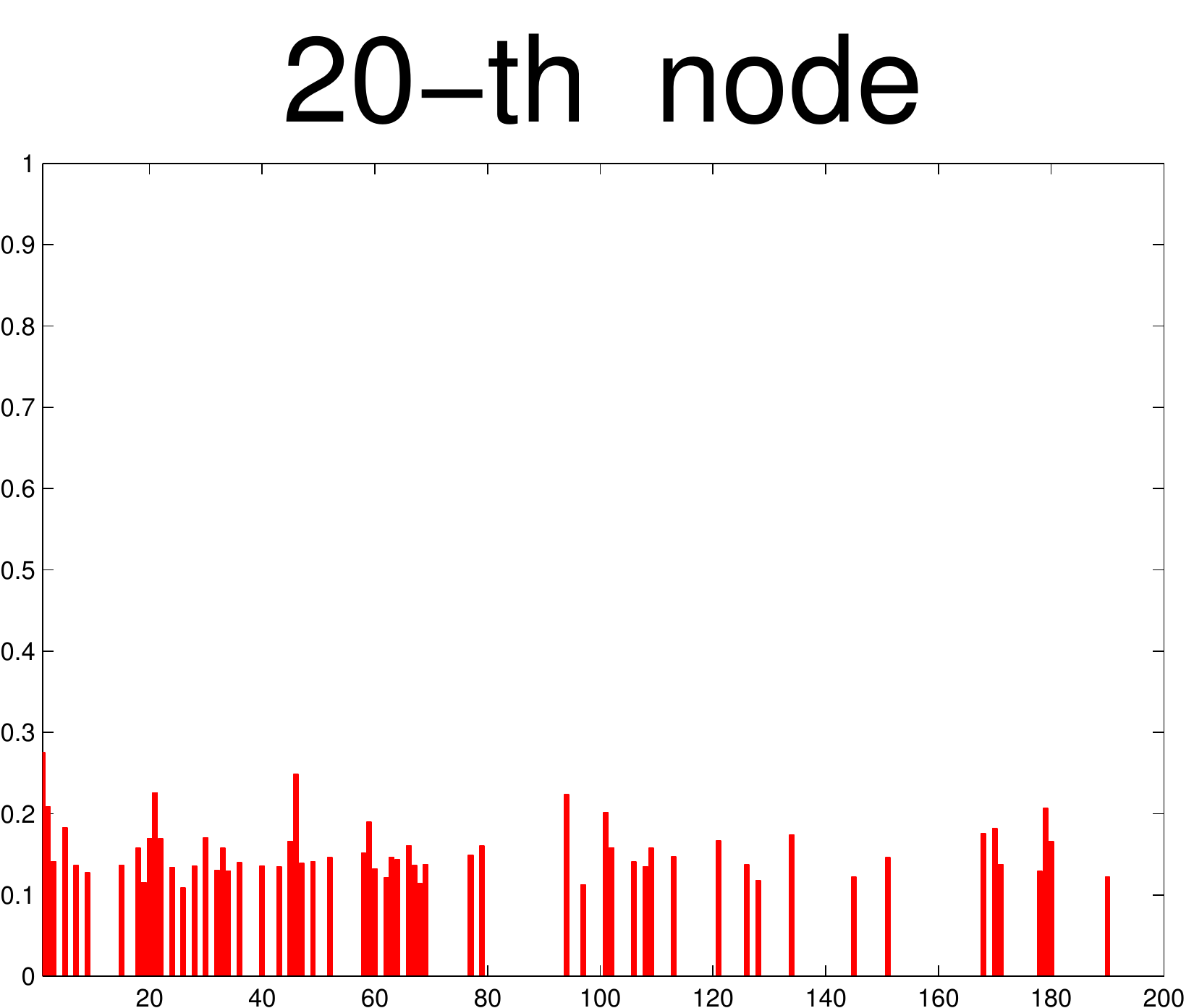}
    \end{center}
    \caption{Weak classifiers' coefficients at each
    node. The x-axis represents the index of all $200$ candidate
    weak classifiers.
    }
    \label{fig:fddb_rmfeat}
  \end{figure}

  \begin{figure}[t]
    \begin{center}
    \subfigure[]{
        \includegraphics[width=0.37\textwidth,clip]{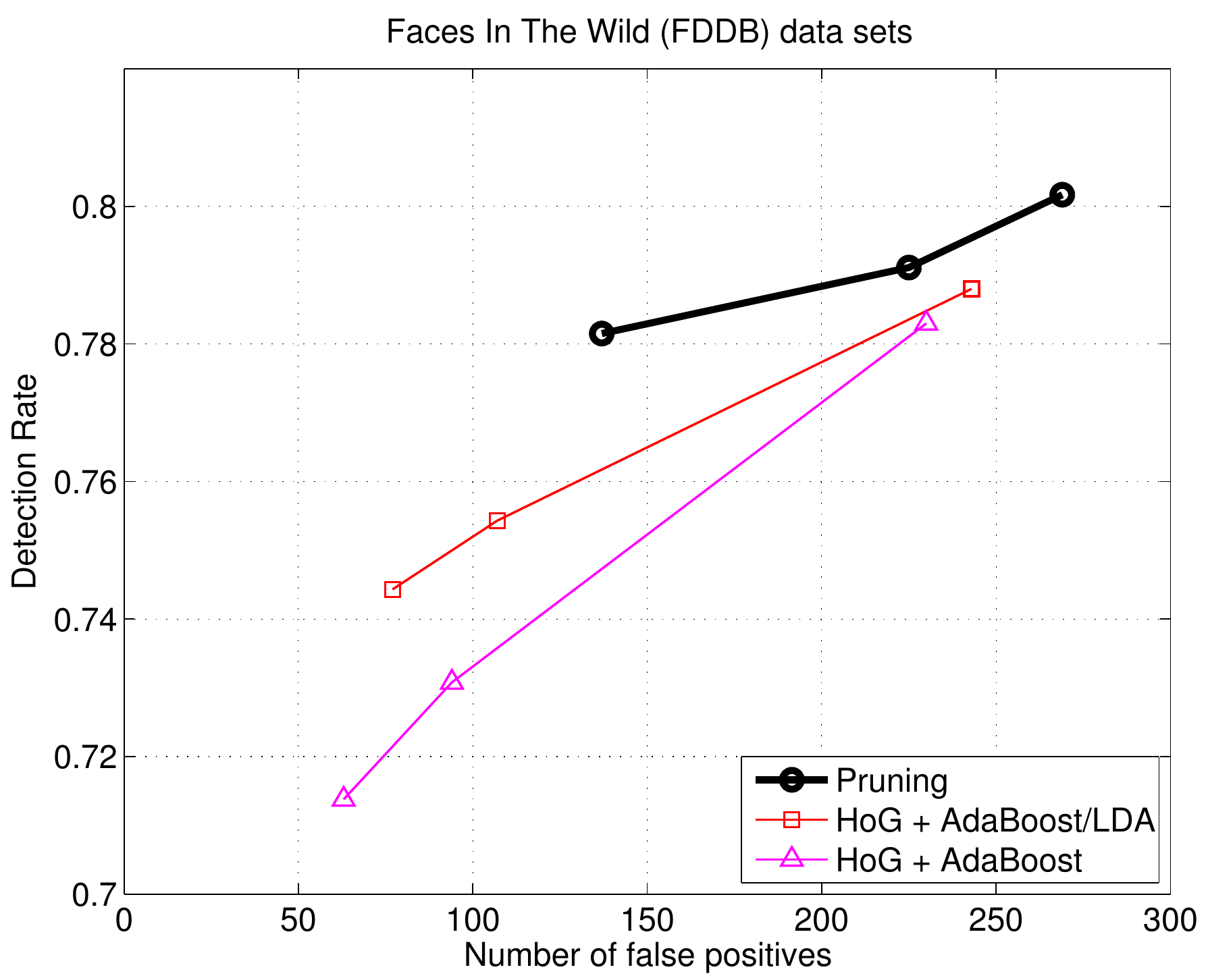}
        \label{fig:fig2a}
    }
    \subfigure[]{
        \includegraphics[width=0.357\textwidth,clip]{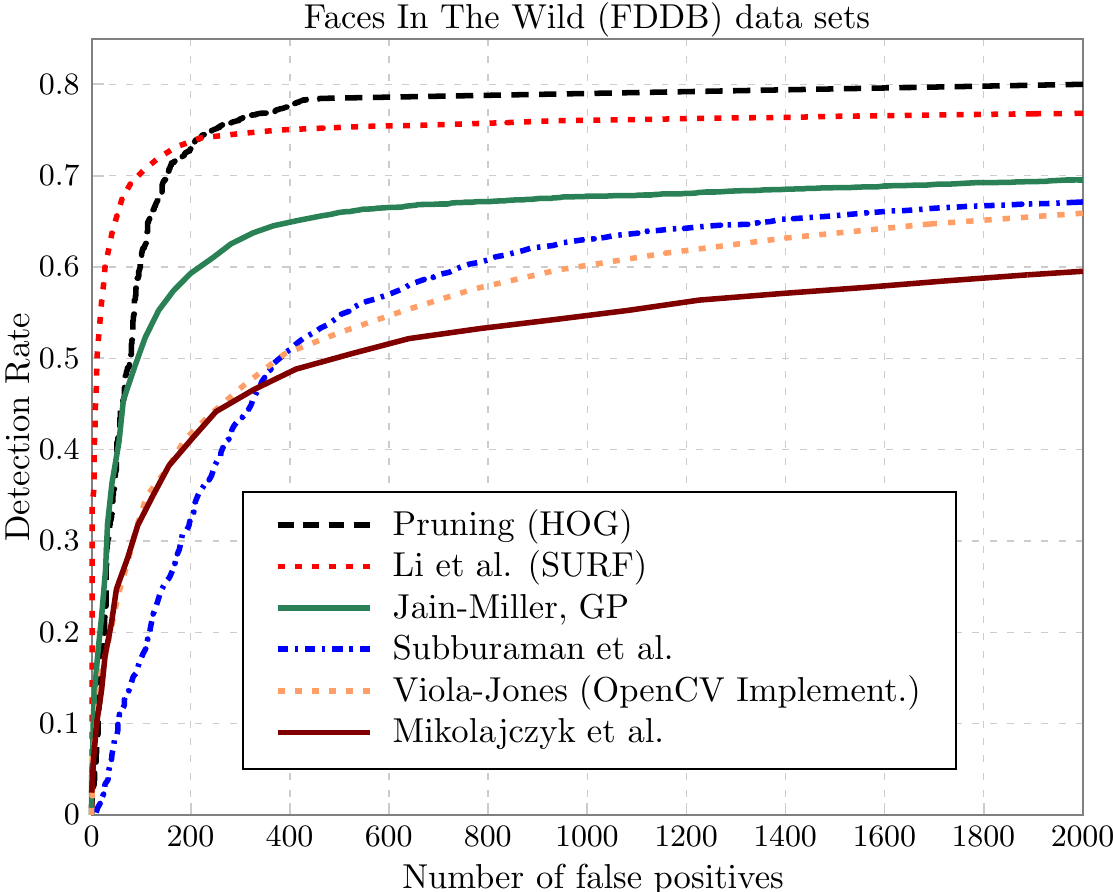}
        \label{fig:fig2b}
    }
    \end{center}
    \caption{Cascade performance on FDDB data sets.
    (a) The curve is generated by adding one cascade level at a time.
    (b) The curve is generated by using the FDDB evaluation code
    \cite{Jain2010FDDB}.
    The results are the average of $10$ splits.
    }
    \label{fig:fig2}
  \end{figure}

  \begin{figure*}[t]
    \begin{center}
        \includegraphics[width=0.7\textwidth,clip]{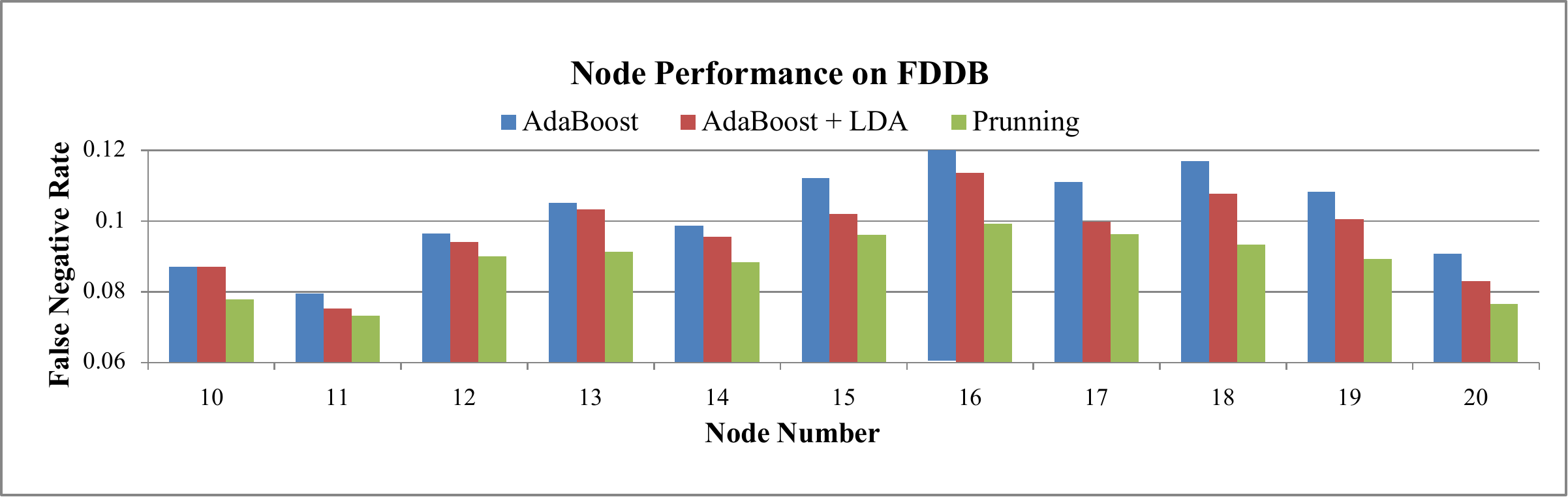}
    \end{center}
    \caption{Average independent node performance on FDDB face detection test sets.
    The results are the average of $10$ splits.
    }
    \label{fig:fig2node}
  \end{figure*}

  \begin{figure}[t]
    \centering
        \includegraphics[width=0.12\textwidth,clip]{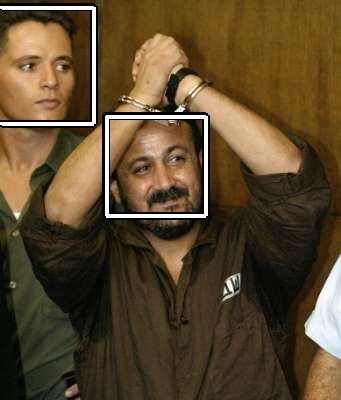}
        \includegraphics[width=0.12\textwidth,clip]{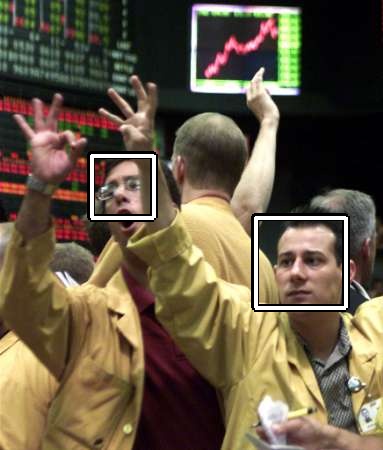}
        \includegraphics[width=0.12\textwidth,clip]{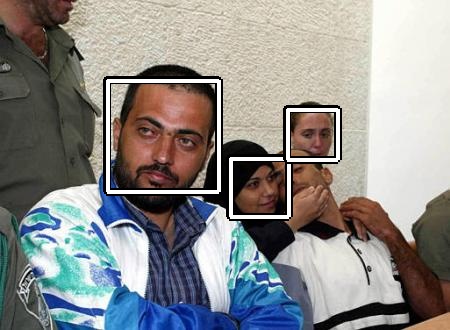}
        \includegraphics[width=0.12\textwidth,clip]{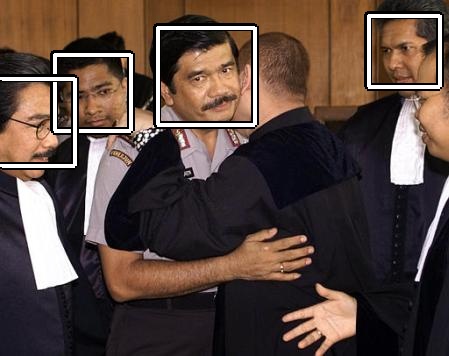}
        \includegraphics[width=0.12\textwidth,clip]{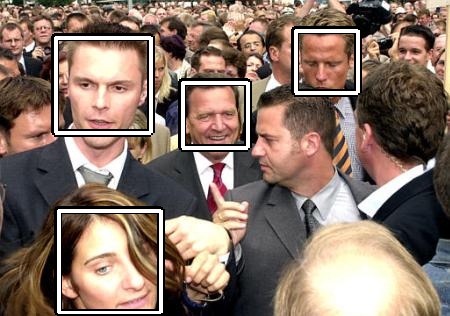}
        \includegraphics[width=0.12\textwidth,clip]{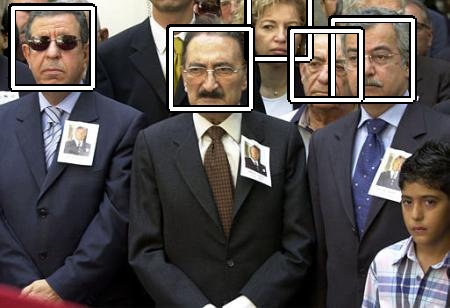}
    \caption{Detection examples on challenging FDDB test sets.
    Despite a large amount of variation in poses and self occlusion,
    our system is still able to perform exceptionally well.
    }
    \label{fig:examples_FDDB}
  \end{figure}

  \begin{figure}[t]
    \begin{center}
        \includegraphics[width=0.15\textwidth,clip]{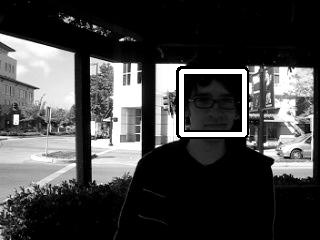}
        \includegraphics[width=0.15\textwidth,clip]{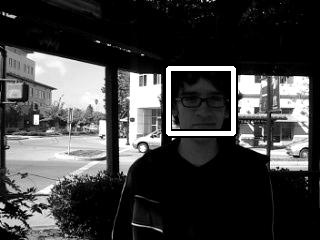}
        \includegraphics[width=0.15\textwidth,clip]{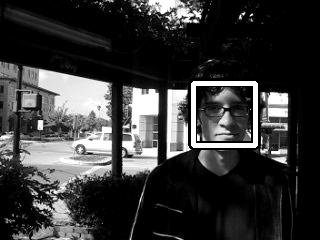}
        \includegraphics[width=0.15\textwidth,clip]{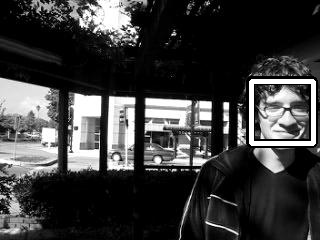}
        \includegraphics[width=0.15\textwidth,clip]{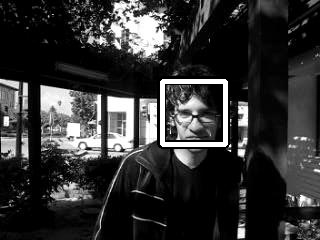}
        \includegraphics[width=0.15\textwidth,clip]{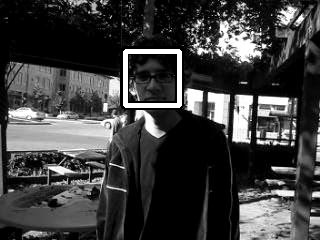}
    \end{center}
    \caption{Detection examples on Trellis data sets \cite{Ross2004Adaptive}.
    The video contains $501$ frames of a person moving underneath a trellis with large illumination and pose changes.
    On this video, our algorithm runs at $5$ frames per second using HOG features.
    }
    \label{fig:examples_david}
  \end{figure}

    In this experiment, we evaluate our algorithm on more challenging face detection
    benchmark, FDDB database \cite{Jain2010FDDB}.  The data sets consist of faces under varying
    conditions in unconstrained environments.  The database consists of $5,171$ faces in $2,845$
    images.  Similar to the experiment on single-node detectors, we use
    HOG features to train a face detector.  A cascade of $20$ nodes ($545$ weak
    classifiers) is trained.
    The number of weak classifiers in each node is $4$, $7$, $10$, $10$,
    $12$, $12$, $15$, $15$, $20$, $20$, $30$, $30$, $30$,
    $40$, $40$, $40$, $50$, $50$, $50$, $60$.
    To ensure a fair comparison, we used the same cascade structure and
    the same number of weak classifiers for all learning methods.
    During evaluation, we set the scale
    factor to $1.25$ and the stride step to $4$ pixels.

    Since our detector is based on the cascade classifier framework,
    we evaluate our algorithms only on discrete score approach \cite{Jain2010FDDB}.
    In this scheme, each detection is evaluated either as a match or non-match based on the ratio of
    overlapping areas.  To be considered a correct detection, the area of overlap between the
    predicted bounding box and the ground truth bounding box must exceed $50\%$, using
    the PASCAL object detection criterion.  Post-processing
    similar to \cite{Viola2004Robust} is applied to merge final detection results.
    We report our results based on the average of $10$ split runs
    (the FDDB database provides $10$ groups of face data).
    At each run, we generate the positive training data by combining
    faces from $9$ groups and use the face data in the remaining group for testing.
    We also mirror the positive training data.
    In total, there are approximately $8,000$ training faces.
    For our pruning approach, we set $T_1$ to $200$.

    Fig.~\ref{fig:fig2a} shows ROC curves of our approach, AdaBoost and AdaBoost+LDA.
    In this experiment, we remove
    faces which have a resolution smaller than $48 \times 48$ pixels from our test sets.  The curve
    is generated by adding one cascade level at a time.  For example, the rightmost marker
    corresponds to the detection rate and the number of false positives using $20$ levels of cascade
    classifiers.  Similar to the previous experiment, pruning performs better than other approaches.
    We also plot the value of weak classifiers' coefficients of
    our approach
    at the $2$-nd, $4$-th, $8$-th, $12$-th, $16$-th and $20$-th node
    in Fig.~\ref{fig:fddb_rmfeat}.
    The $x$-axis represents the index of all $200$ candidate weak classifiers
    and the $y$-axis represents weights of weak classifiers.
    We observe that our approach selects similar weak classifiers as AdaBoost
    in early stages of the cascade (a large number of non-zero coefficients
    appear close  the first few indices in the second and fourth node).
    However, in later stages, the set of selected weak classifiers tends
    to spread out across $200$ candidate weak classifiers.
    We also observe that weak learners' coefficients in later stages are more
    uniformly distributed.
    We suspect that a large number of negative training samples in later
    stages consist of difficult-to-classify background patches,
    \ie, the bootstrapping process has discarded a lot of easy-to-classify negative samples.
    As a result, weak learners'
    coefficients are more uniformly distributed as no single weak classifier
    can perform better than other weak classifiers.

    Fig.~\ref{fig:fig2b} compares ROC curves of various face detectors.
    We use the FDDB
    evaluation software and their original ground-truth data to generate performance curves.
    The evaluation program requires detection result's
    coordinates, width, height
    and the confidence score associated with the detection window.
    In our experiment, we merge multiple detection windows into a single
    detection window and calculate their average detection window positions.
    The confidence score is calculated from the average detection
    responses in the last node, \ie,
    $\sum_{j=1}^T \alpha_j h_j(\cdot)$ where $\{h_j(\cdot)\}_{j=1}^T$
    and $\{\alpha_j\}_{j=1}^T$ are the set of weak classifiers and
    their coefficients in the last node.
    As the baseline, we use
    the face detector of Li \etal \cite{Li2011Face},
    the online approach using Gaussian process regression scheme (VJGPR) \cite{Jain2011Online},
    the OpenCV implementation of Viola and Jones' face detector \cite{Viola2004Robust},
    the probabilistic based part detector \cite{Mikolajczyk2004Human} and
    the bounding box estimation based face
    detector \cite{Venkatesh2010Fast}.
    From Fig.~\ref{fig:fig2b}, our detector achieves a comparable performance
    to the SURF face detector of Li \etal \cite{Li2011Face}.
    However, our approach has a better detection rate when the number of false
    positives is greater than $200$.
    At a small number of false positives, we observe that our system performs
    slightly worse than \cite{Li2011Face}.
    We suspect that this is related to how the confidence score is generated.
    \cite{Li2011Face} defines the confidence score as the sum of
    the detection probability of all nodes while our approach uses the confidence score
    from the last node.
    We suspect that our performance at the low false positive rate
    can be further improved by exploiting weak classifiers
    learned in previous nodes (similar to the soft
    cascade \cite{Bourdev2005Robust} and the multi-exit
    classifier \cite{PhamHC08}).
    In addition, training another set of coefficients,
    which maximizes the area under the ROC curve (AUC), would be
    a better alternative in order to generate better confidence scores.
    We leave this as a future work.

    Fig.~\ref{fig:fig2node} shows an average independent node comparison on FDDB test sets of
    different detectors.
    We observe that pruning has a smaller false negative rate than AdaBoost and AdaBoost+LDA in every
    node.
    We observe that the improvement of our approach over AdaBoost+LDA at each node
    is more consistent in Fig.~\ref{fig:fig2node} than in Fig.~\ref{fig:mitcmu_node}.
    We suspect that this may be due
    to:  (a) the use of $10$ split runs,
    (b) FDDB database is much more challenging than the
    MIT-CMU test set and, hence, there is more room for performance improvement.
    Again, our results demonstrate the
    evidence that pruning can further enhance the final performance.
    Table~\ref{tab:traintime}
    shows the approximate cascade training time and the average number of features evaluated per
    patch.
    Note that the training time of HOG features is faster than the training time of
    Haar-like features due to the use of OpenMP for parallelized feature extraction and weak
    classifier training.
    It is important to note here the difference between Fig.~\ref{fig:singlenode}
    and Table~\ref{tab:traintime}.
    Fig.~\ref{fig:singlenode} shows the detection performance of
    a single-node classifier while
    Table~\ref{tab:traintime} shows the average number of features evaluated per
    image patch during testing for the cascade classifier of $20$ nodes.
    The threshold in Fig.~\ref{fig:singlenode} was chosen such that
    half non-faces {\em in the test set} will be correctly classified
    and the other half will be incorrectly classified
    (a false positive rate of $50\%$ on the test set).
    From that experiment, we observe that our approach has
    a slightly higher detection rate than AdaBoost+LDA when
    both algorithms have the same number of weak classifiers.
    On the other hand, the threshold in Table~\ref{tab:traintime} was
    chosen such that half non-faces {\em in the training set of that node}
    will be passed to the next node classifier and
    the other half will be discarded (a false positive
    rate of $50\%$ on the training set of the given node).
    From Table~\ref{tab:traintime}, we observe that each node
    of our approach removes less background patches than AdaBoost+LDA,
    \ie, a higher average number of features evaluated per patch.
    In contrast, both AdaBoost and AdaBoost+LDA have a very similar
    average number of features evaluated per image patch.
    Our conjecture is that AdaBoost+LDA uses the same set of
    features as AdaBoost, therefore,
    it has a similar true negative rate as AdaBoost.
    In contrast, our approach selects a set of features based on
    the asymmetric node learning criterion.
    This often leads to a better generalization performance than a cascade of
    AdaBoost+LDA but it can lead to a slightly higher number of evaluated
    features during test time.
    Fig.~\ref{fig:examples_FDDB} shows some detection examples on challenging
    FDDB data sets.

    In the next experiment, we use a test video  ``Trellis''
    obtained from
    \cite{Ross2004Adaptive} to test the real-time performance of different detectors.  The video
    contains $501$ frames of $320 \times 240$ pixels images.  It contains a person moving underneath
    a trellis with large illumination changes and pose variations.  We evaluate all $10$ detectors
    obtained from the previous experiment.  During evaluation, the video
    is scanned with a stride of $4$ pixels in both directions.  We apply a scale ratio of $1.25$ for
    multiple scale detection.  We evaluate our detection using PASCAL criteria, \ie, the overlapping
    ratio between the detection and the ground truth must exceed $50\%$ to be considered as the correct
    detection.  We record the number of false positives, true detections and evaluation time
    obtained from $20$ levels of cascade classifiers.  Table~\ref{tab:david} shows our results by
    averaging the number of false positives, detection rate and evaluation time.  We test our
    detectors using a standard desktop computer (Intel core i-$7$ CPU $930$ with $12$ GB memory).
    Fig.~\ref{fig:examples_david} shows some detection examples of our approach.

  \begin{table}[tb!]
  \caption{
  CPU time performance and detection rates
  of different detectors on the ``Trellis'' video sequence
  ($320 \times 240$ pixels resolution).
  The same number of weak classifiers and nodes are used in all detectors.
  }
  \begin{center}
    \begin{tabular}{l|c|c|c}
    \hline
    Algorithm & FPS & Det. rate & $\#$ false pos.$/$image \\
    \hline
    \hline
      Pruning & $5.0$ & $26.0\%$ & $0.0016$ \\
      AdaBoost \cite{Viola2004Robust} &	$5.2$ & $16.3\%$ & $0.0002$ \\
      AdaBoost+LDA \cite{Wu2008Fast}& $5.0$ & $22.2\%$ & $0.0014$ \\
     \hline
    \end{tabular}
  \end{center}
  \label{tab:david}
  \end{table}

\subsection{Car Detection}

  \begin{figure}[t]
    \begin{center}
        \includegraphics[width=0.11\textwidth,clip]{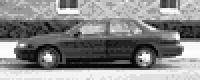}
        \includegraphics[width=0.11\textwidth,clip]{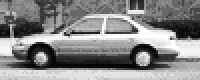}
        \includegraphics[width=0.11\textwidth,clip]{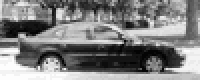}
        \includegraphics[width=0.11\textwidth,clip]{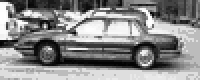}
        \includegraphics[width=0.11\textwidth,clip]{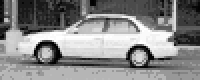}
        \includegraphics[width=0.11\textwidth,clip]{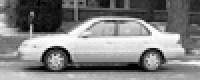}
        \includegraphics[width=0.11\textwidth,clip]{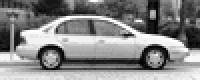}
        \includegraphics[width=0.11\textwidth,clip]{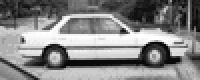}
        \includegraphics[width=0.11\textwidth,clip]{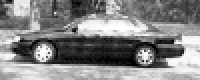}
        \includegraphics[width=0.11\textwidth,clip]{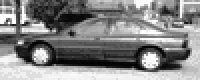}
        \includegraphics[width=0.11\textwidth,clip]{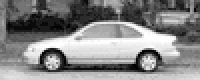}
        \includegraphics[width=0.11\textwidth,clip]{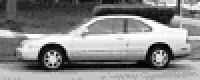}
    \end{center}
    \caption{A sample of UIUC car training images.
    }
    \label{fig:car_train_images}
  \end{figure}

   \begin{figure}[t]
    \centering
        \includegraphics[width=0.35\textwidth,clip]{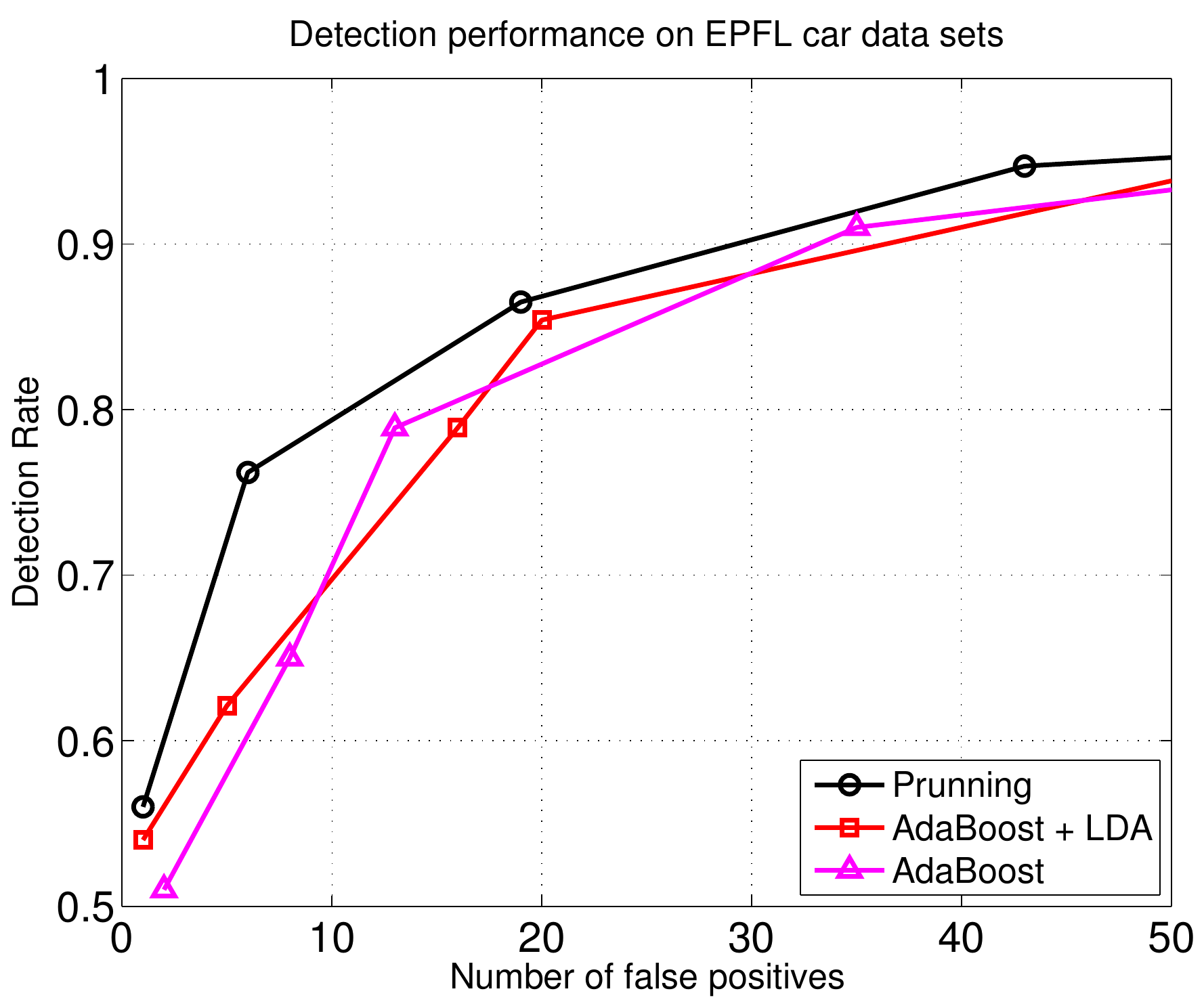}
    \caption{Performance on $512$ images of side view cars from EPFL data sets \cite{Ozuysal2009Pose}.
    }
    \label{fig:UIUC_EPFL}
  \end{figure}

  \begin{figure}[b!]
  \label{fig:UIUC_EPFL2}
  \centering
        \includegraphics[width=0.125\textwidth,clip]{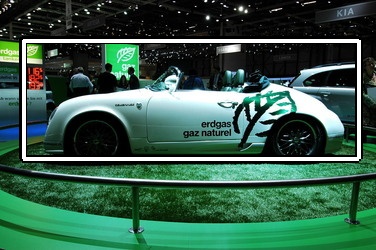}
        \includegraphics[width=0.125\textwidth,clip]{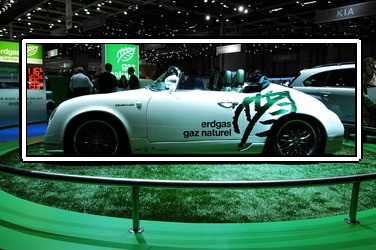}
        \includegraphics[width=0.125\textwidth,clip]{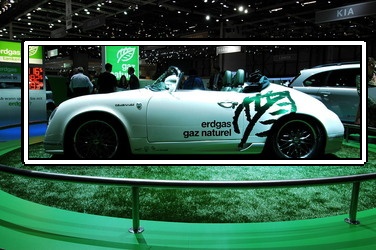}
        \\
        \includegraphics[width=0.125\textwidth,clip]{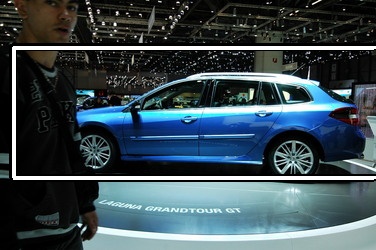}
        \includegraphics[width=0.125\textwidth,clip]{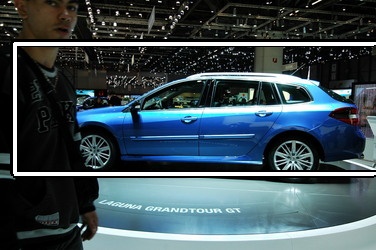}
        \includegraphics[width=0.125\textwidth,clip]{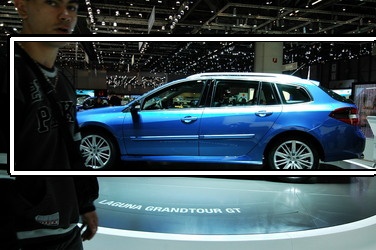}
        \\
        \includegraphics[width=0.125\textwidth,clip]{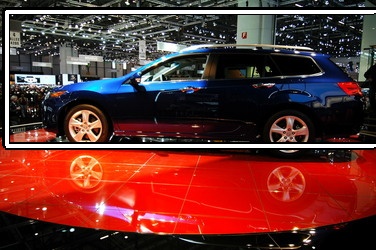}
        \includegraphics[width=0.125\textwidth,clip]{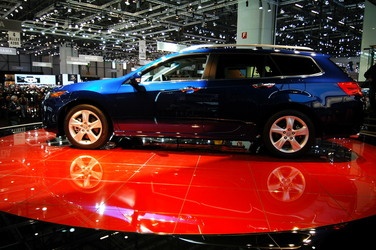}
        \includegraphics[width=0.125\textwidth,clip]{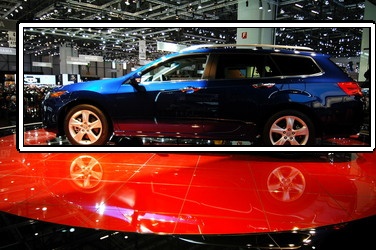}
        \\
        \includegraphics[width=0.125\textwidth,clip]{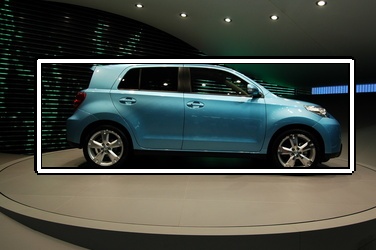}
        \includegraphics[width=0.125\textwidth,clip]{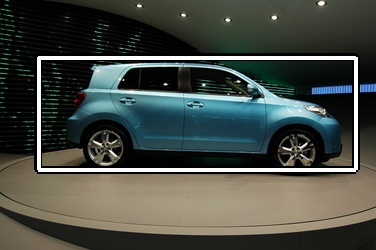}
        \includegraphics[width=0.125\textwidth,clip]{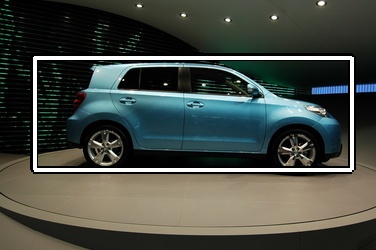}
        \\
        \includegraphics[width=0.125\textwidth,clip]{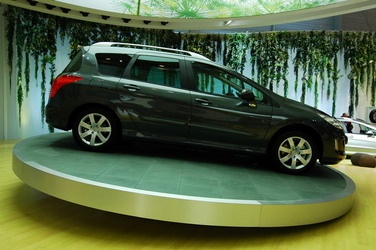}
        \includegraphics[width=0.125\textwidth,clip]{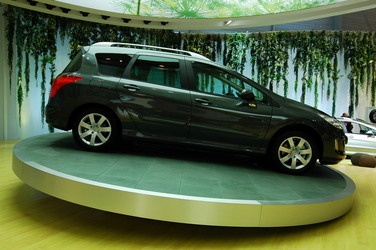}
        \includegraphics[width=0.125\textwidth,clip]{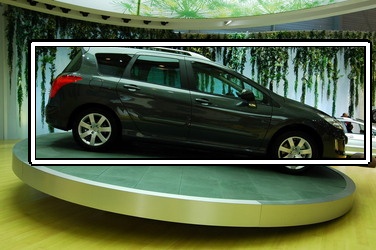}
        \\
        \includegraphics[width=0.125\textwidth,clip]{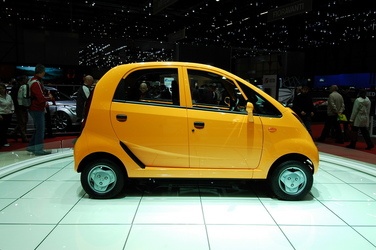}
        \includegraphics[width=0.125\textwidth,clip]{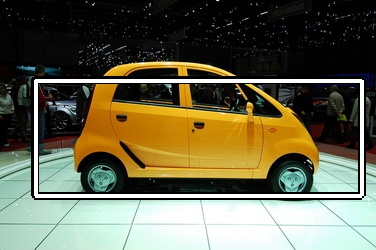}
        \includegraphics[width=0.125\textwidth,clip]{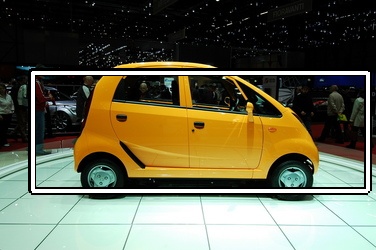}
        \\
        \includegraphics[width=0.125\textwidth,clip]{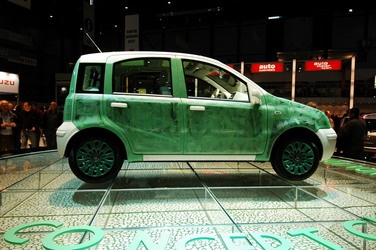}
        \includegraphics[width=0.125\textwidth,clip]{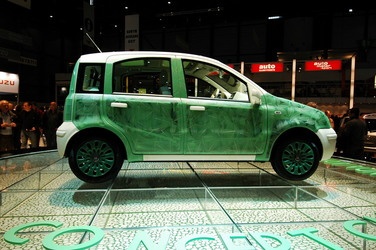}
        \includegraphics[width=0.125\textwidth,clip]{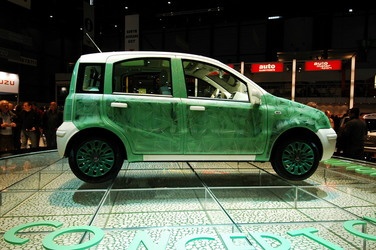}
    \caption{Detection results from the car show environment using
    AdaBoost (first column), AdaBoost+LDA (middle column) and our
    pruning approach (last column).  Test data sets contain cars of
    type sports sedans, hatchbacks and station wagons.  These car
    models are different from those used during training (mostly
    typical sedans) (Fig.\ \ref{fig:car_train_images}).  Despite these
    differences, our detectors detect most cars correctly.
    }
  \end{figure}

\begin{table}[t!]
  \caption{Performance on UIUC multi-scale car test sets}
  \begin{center}
    \begin{tabular}{l|c|c|c}
    \hline
    Algorithm & F-Measure & Det. rate & $\#$ false pos. \\
    \hline
    \hline
      Pruning & \textbf{$98.6\%$} & $97.8\%$ & $1$ \\
      AdaBoost \cite{Viola2004Robust} &	\textbf{$98.6\%$} & $98.6\%$ & $2$ \\
      AdaBoost+LDA \cite{Wu2008Fast}& \textbf{$98.6\%$} & $97.8\%$ & $1$ \\
      CS-AdaBoost \cite{MasnadiShirazi2011Cost} & $95.26\%$ & $95.5\%$ & $9$ \\
      Agarwal \etal \cite{Agarwal2004Learning} & $43.4\%$ & $38.9\%$ & $56$ \\
     \hline
    \end{tabular}
  \end{center}
  \label{tab:UIUC1}
\end{table}

  \begin{table}[t!]
  \caption{Average number of HOG features evaluated per patch.}
  \begin{center}
    \begin{tabular}{l|c|c}
    \hline
    Algorithm & UIUC test sets & EPFL test sets \\
    \hline
    \hline
      Pruning & $22.6$ & $20.3$ \\
      AdaBoost \cite{Viola2004Robust} &	$24.6$ & $23.1$ \\
      AdaBoost+LDA \cite{Wu2008Fast}& $24.6$ & $23.1$  \\
     \hline
    \end{tabular}
  \end{center}
  \label{tab:UIUC_hog}
  \end{table}

    Next, we conduct an experiment on car detection.
    We compare different detectors on the UIUC data sets \cite{Agarwal2004Learning}.
    Training sets consist of $550$ mixed left and right profile views of cars with a resolution $40 \times 100$ pixels.
    Fig.~\ref{fig:car_train_images} shows some random samples of UIUC car training images.
    We combine both left and right profile views, with their mirrored
    samples and train a single detector.
    Our training sets consist of $1,100$ car samples.
    Negative training images used are the same as those used in face experiments.
    The detector is evaluated on $108$ multi-scale test images consisting of $139$ cars.

    We compare our pruning approach to AdaBoost and AdaBoost+LDA.
    We evaluate our algorithm on HOG features.
    We define blocks with following scales (minimum of $4 \times 4$ pixels and maximum of $40 \times 100$ pixels) and width-length ratios of $1:1$, $1:2$, $2:1$, $1:3$ and $3:1$.
    Each block is divided into $2 \times 2$ cells, and HOG in each cell is divided into $9$ bins.
    There are a total of $3,801$ blocks.
    An $\ell_1$-Sqrt normalization is applied to the feature vector \cite{Dalal2005Histograms}.
    At each iteration, we randomly sample $25\%$ of all possible blocks for training a weak classifier.
    We train a visual detector of $20$ cascade nodes.
    The number of weak classifiers in each node is $4$, $7$, $10$, $10$,
    $12$, $12$, $15$, $15$, $20$, $20$, $30$, $30$, $30$,
    $40$, $40$, $40$, $50$, $50$, $50$, $60$.
    For our pruning approach, we set $T_1$ to $300$.
    We follow the technique used in \cite{Viola2004Robust} to merge multiple detection windows.

    To evaluate our performance, we use the software provided along with the data set.
    An output is counted as a correct detection if it lies within $25\%$ of the true object dimension in each direction.
    Only one detection window is counted as correct if two or more detection windows satisfy the criteria.
    We record the performance by F-measure, detection rate and the number of false detections.
    The F-measure is the weighted harmonic mean of precision and recall.
    Table~\ref{tab:UIUC1} shows the performance of different detectors.
    We provide a baseline system of \cite{Agarwal2004Learning} and \cite{MasnadiShirazi2011Cost}.
    Note that the method of \cite{Agarwal2004Learning} and \cite{MasnadiShirazi2011Cost} were trained with only $500$ negative patches so it can not be directly compared with our algorithms.
    Also, \cite{MasnadiShirazi2011Cost} uses a single-scale test set.
    From the table, our evaluated algorithms achieve similar F-Measure and perform similarly on UIUC multi-scale test sets.
    In the next experiment, we test our detectors on more challenging EPFL car
    data sets {Ozuysal2009Pose}.
    We use $512$ images of left and right profile poses from $20$ car models.
    We evaluate our detection results against provided ground truths.
    We use PASCAL criteria for this experiment.
    Fig.~\ref{fig:UIUC_EPFL} compares ROC curves of different approaches.
    Our results clearly indicate that pruning can further improve the generalization performance of visual detectors.
    Fig.~\ref{fig:UIUC_EPFL2} demonstrates some car detection results using different algorithms.
    It is quite interesting to observe that our detector is able to detect car in the penultimate row but not in the last row.
    We suspect our detector fails to detect a car whose boot has a vertical shape.
    This is not surprising since this particular shape is not common in our training samples.
    We compare the average number of HOG features used in Table~\ref{tab:UIUC_hog}.
    For car data sets, the proposed
    pruning not only gives a higher detection rate but also
    performs faster than other evaluated algorithms
    (as it requires less number of HOG features on average).

\subsection{Discussion}
In this section, we briefly discuss our experimental results reported
earlier and two important assumptions we made in our experiments.
In the previous section, we evaluate our detector on two challenging
object types, namely faces in unconstrained environments and
unconstrained vehicle types.
We observe that our approach outperforms other algorithms on both
face and car data sets (Fig.~\ref{fig:mitcmu_node}, \ref{fig:fig2node},
\ref{fig:UIUC_EPFL} and Table~\ref{tab:UIUC1}).
In terms of evaluation time, we observe that our approach is slightly faster
than both AdaBoost and AdaBoost+LDA on car data sets (Table~\ref{tab:UIUC_hog})
but slightly slower than both AdaBoost and AdaBoost+LDA
on face data sets (Table~\ref{tab:traintime}).
As previously discussed, both AdaBoost and AdaBoost+LDA have a
similar average number of features evaluated per image patch.
In contrast, our approach selects a set of features based on
the asymmetric node learning criterion.
These selected features often leads to a better generalization performance
compared to those used in AdaBoost+LDA.
However, they can result in a slightly less number of average
features that needs to be extracted during testing (for car data sets)
or a slightly higher number of average
features per image patch during testing (for face data sets).

It is important to note there that the success of
our approach is based on two key assumptions.
The first assumption is that the set
of weak learners selected must be normally distributed.
For object detection problems, Wu \etal demonstrate that
this assumption is often valid in practice, \ie,
$\bw^\T \bh$, is approximately Gaussian for most
instantiations of $\bh$ \cite{Wu2008Fast}.
Here $\bh = [h_1(\cdot), h_2(\cdot), \cdots,
h_T(\cdot)]$.
Setting $T_1$ to be large in our experiments ensures that
the initial set of features is normally distributed as
the mean of a sufficiently large number of independent random
variables will be approximately
normally distributed (the central limit theorem).
The second assumption is that the covariance matrix, $S_w$, is non-singular.
In other words, the approach fails when the covariance
matrix does not have a full rank and its inverse does not exist.
In order to avoid this ill-posed problem, we can regularize the matrix
$S_w$ by $S_w = S_w + \lambda \I$, where $\I$ is the identity matrix and
$\lambda$ is the regularization parameter.
This technique is also known as Tikhonov regularization.

\section{Conclusion}
\label{sec:conc}

We have presented a two-stage approach for training a visual object detector by separating the
feature extraction process from constructing the asymmetric classifier.
The learned ensemble better meets the asymmetric node learning and consequently
improves the detection performance of the entire cascade.
Experiments on various data sets show that
the proposed method consistently outperforms
the traditional framework of Viola-Jones \cite{Viola2004Robust}
as well as the LDA post-processing algorithm of Wu \etal \cite{Wu2008Fast}.
We have also demonstrated empirically that training the HOG cascade
classifier using our approach is as competitive as state-of-the-art methods in the literature.
On
FDDB data sets, our approach substantially outperforms all other methods evaluated and achieves the
state-of-the-art performance.  In the future, we plan to investigate asymmetric learning objectives
in multi-class problems.  Another direction of future research would be to investigate whether
parameter $\gamma$, in \eqref{EQ:within_class_cov}, should be adjusted in each cascade node.

\bibliographystyle{ieee}
\bibliography{gslda}

\end{document}